\documentclass{article}

\usepackage{xcolor}

\usepackage[preprint]{corl_2026} %

\usepackage[numbers]{natbib}
\usepackage{multicol}
\usepackage{multirow}

\usepackage{amsmath,amssymb,amsfonts}
\usepackage{algorithmic}
\usepackage{graphicx}
\usepackage{textcomp}
\usepackage{booktabs}
\usepackage{tabularx}
\usepackage{amsmath}
\usepackage{amssymb}
\usepackage{amsfonts}
\usepackage{bm}
\usepackage{makecell}
\usepackage{comment}
\usepackage{siunitx}
\usepackage{float}
\usepackage[normalem]{ulem}
\usepackage{subcaption}
\usepackage{caption}
\usepackage{placeins}
\usepackage{enumitem}
\usepackage{wrapfig}
\usepackage{arydshln}
\usepackage{xspace}
\newcommand{\ourbenchmark}{AM-Bench\xspace}
\newlength{\authorblocktabcolsep}

\title{AM-Bench: A Modular Simulation Suite and Benchmark for Aerial Manipulation Policy Learning}

\hypersetup{
  pdftitle={AM-Bench: A Modular Simulation Suite and Benchmark for Aerial Manipulation Policy Learning},
  pdfauthor={Yutong Wang; Dongjae Lee; Xiaofeng Guo; Yuanzhu Zhan; Yufei Jiang; Bavin Saravanan; Muqing Cao; Jia Xie; Chenyang Mao; Sebastian Scherer; Junyi Geng; Guanya Shi}
}

\author{
  {\footnotesize
  \textbf{Yutong Wang\textsuperscript{1,*}}\hspace{0.6em}
  \textbf{Dongjae Lee\textsuperscript{1,3,*}}\hspace{0.6em}
  \textbf{Xiaofeng Guo\textsuperscript{1,*}}\hspace{0.6em}
  \textbf{Yuanzhu Zhan\textsuperscript{2}}\hspace{0.6em}
  \textbf{Yufei Jiang\textsuperscript{2}}\hspace{0.6em}
  \textbf{Bavin Saravanan\textsuperscript{1}}}\\[0.15em]
  {\footnotesize
  \textbf{Muqing Cao\textsuperscript{1}}\hspace{0.6em}
  \textbf{Jia Xie\textsuperscript{1}}\hspace{0.6em}
  \textbf{Chenyang Mao\textsuperscript{1}}\hspace{0.6em}
  \textbf{Sebastian Scherer\textsuperscript{1}}\hspace{0.6em}
  \textbf{Junyi Geng\textsuperscript{2}}\hspace{0.6em}
  \textbf{Guanya Shi\textsuperscript{1}}}\\[0.35em]
  {\scriptsize\normalfont
  \textsuperscript{1}The Robotics Institute, School of Computer Science,
  Carnegie Mellon University}\\[-0.15em]
  {\scriptsize\normalfont
  \textsuperscript{2}Department of Aerospace Engineering, Pennsylvania State University}\\[-0.15em]
  {\scriptsize\normalfont
  \textsuperscript{3}School of Mechanical Engineering, Kyung Hee University}\\[-0.05em]
  {\scriptsize\normalfont\textsuperscript{*}Equal contribution.}
}

\begin{document}
\setlength{\authorblocktabcolsep}{\tabcolsep}
\setlength{\tabcolsep}{0pt}
\maketitle
\setlength{\tabcolsep}{\authorblocktabcolsep}

\vspace{-0.8cm}

\begin{figure}[h]
  \centering
  \includegraphics[width=0.99\linewidth,trim=0 3.5cm 0 0,clip]{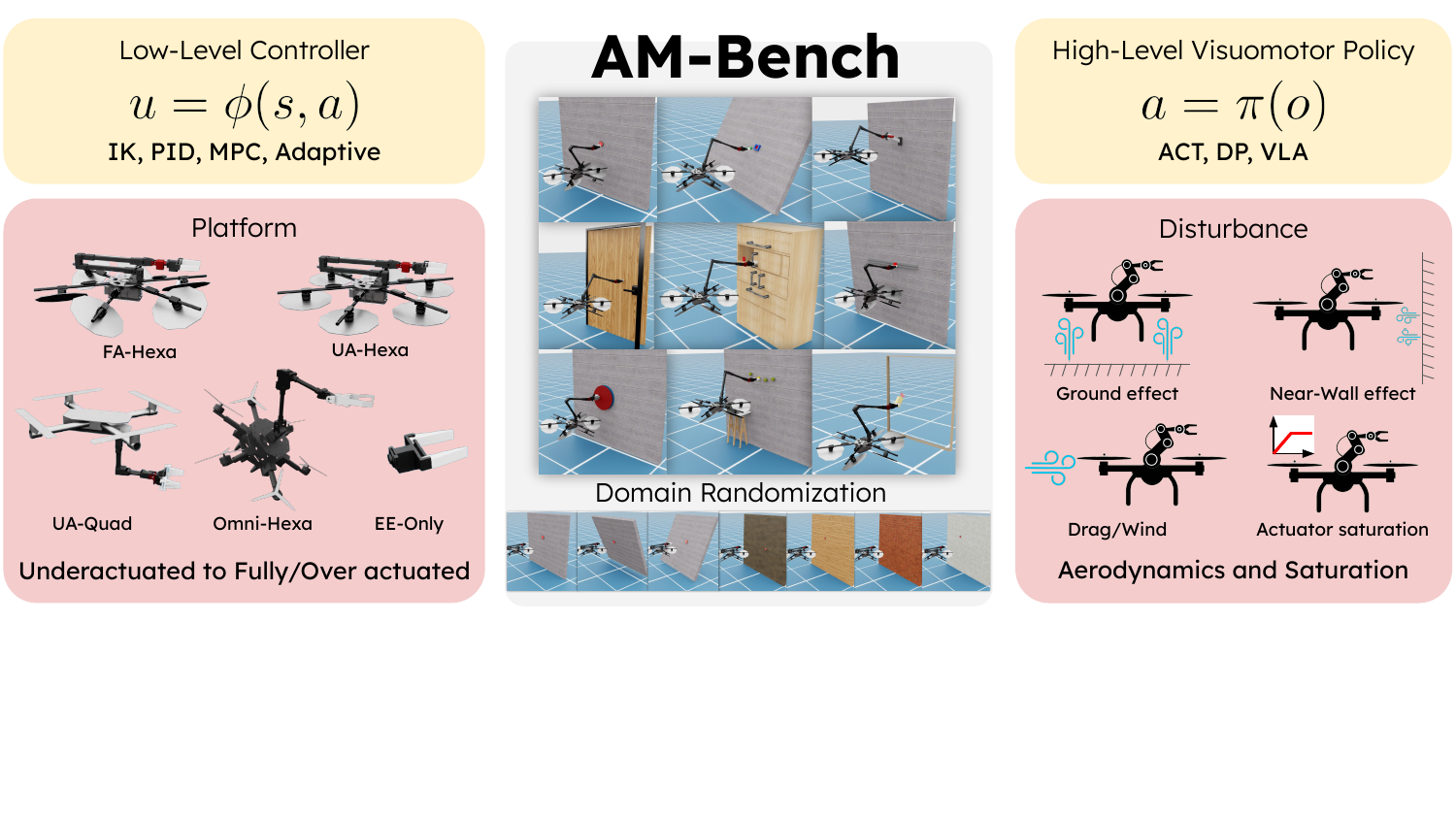}
  \caption{\textbf{Overview of AM-Bench}: AM-Bench is designed to systematically evaluate aerial manipulation along several dimensions: (left) robot platforms, including underactuated (UA-Quad, UA-Hexa), fully actuated (FA-Hexa), and overactuated (Omni-Hexa) systems, together with different low-level controllers; (center) a diverse suite of manipulation tasks with domain randomization in spatial poses and textures; and (right) various visuomotor policy architectures and realistic physical constraints, including aerodynamic disturbances (ground and near-wall effects) and actuator saturation. $u$, $\phi(\cdot, \cdot)$, $s$, $a$, $\pi(\cdot)$, and $o$ denote the control input, controller, state, high-level action, policy, and observation, respectively.}
  \label{fig:teaser}
\end{figure}

\vspace{-0.5cm}

\begin{abstract}
Standardized benchmarks have played a central role in advancing robot manipulation learning, yet most focus on ground-supported manipulation systems, which limits their applicability to dynamics-critical domains such as aerial manipulation (AM). AM presents distinct system-level challenges, including environmental disturbances, coupled dynamics between the manipulator and floating base, and constrained degrees of freedom. Consequently, task performance depends jointly on robot embodiment, low-level control, and high-level policy design. We introduce \ourbenchmark, a modular simulation suite and benchmark for multirotor-based AM policy learning. \ourbenchmark includes representative embodiments spanning underactuated, fully actuated, and overactuated systems, $12$ tasks across contact, transport, and constrained interaction, configurable aerodynamic disturbances and actuator saturation, standard low-level controllers, and baseline policy-learning algorithms. Unlike prior manipulation benchmarks that primarily emphasize end-to-end policy performance, \ourbenchmark enables system-level evaluation of how embodiment, control, disturbances, and policy choices interact. We demonstrate its diagnostic value through three simulation studies spanning high-level policies, policy--control interfaces, and embodiments, together with real-world validation of modeled effects and a hardware test of the learning pipeline. The project website is \href{https://ambench.github.io/}{\texttt{https://ambench.github.io/}}.
\end{abstract}


\section{Introduction}
\label{sec: Introduction}

Mobile manipulation has achieved substantial success in recent years, from simple pick-and-place interactions to long-horizon tasks involving both navigation and manipulation in realistic environments~\citep{reister2022combining, ahn2022can, fu2024mobile}. A key driver has been the development of simulation suites and standardized benchmarks, which provide common tasks and evaluation protocols for comparing policy-learning methods in a scalable way~\cite{geng2025roboverse, nasiriany2026robocasa365largescalesimulationframework, yang2026robolabhighfidelitysimulationbenchmark}.
Recently, there has been growing interest in extending manipulation toward more complex and dynamic environments. Aerial manipulation (AM) holds particular promise because aerial robots can interact with environments that are difficult or unsafe to access from the ground, enabling applications such as non-destructive testing (NDT) \citep{bodie2020active}, painting \citep{vempati2018paintcopter, he2023image}, drilling \citep{ding2021design}, and light-bulb installation \citep{gupta2025umi}. Inspired by the success of learning-based methods in mobile manipulation, recent work has begun transferring similar paradigms to AM settings \citep{he2025flying, gupta2025umi}. 

However, existing simulation environments and benchmarks provide limited support for system-level evaluation of policy learning in multirotor-based AM. Aerial manipulation is distinct from tabletop and ground-based mobile manipulation because accurate and robust low-level control remains a major challenge in AM \cite{khamseh2018aerial}. First, AM systems often operate in outdoor, elevated, or near-building scenarios, where disturbances such as wind, aerodynamic effects, and noisy state estimation are unavoidable \cite{zhao2021super}. Second, the coupled dynamics between the floating base and the manipulator further increase system complexity. Finally, these challenges are compounded by strict mechanical and actuation constraints, including limited degrees of freedom, limited workspace and wrench capabilities, actuator saturation, and variation in AM morphologies and actuation designs. As a result, AM policy learning is inherently a system-level problem that requires joint evaluation of embodiment, control, planning, and decision-making rather than isolated components. Given the high costs and safety risks of large-scale real-world experiments, a standardized simulation suite can provide a valuable foundation for fair evaluation and research in AM.

In this work, we present \textbf{\ourbenchmark}, a modular simulation suite and benchmark for policy learning in multirotor-based aerial manipulation. The suite is organized around five main design axes: task environment, robot embodiment, high-level policy interface, low-level control, and physical disturbances and actuator constraints. Its 12 tasks represent recurring real-world AM capabilities in inspection and installation, harvesting and transport, and constrained-contact maintenance---capabilities that remain difficult to execute reliably with today's AM methods. The embodiment axis covers underactuated, fully actuated, and overactuated multirotor-based platforms that have been demonstrated in or are physically motivated by real-world AM systems. 
For high-level policies, the suite exposes common observation, action, and task-conditioning interfaces for evaluating visuomotor policies under controlled settings, including vision-language-action (VLA) models. 
For low-level control, we support standard approaches such as PID, MPC, and $L_1$ adaptive control \cite{wu2025L1quad}.
To enhance physical realism, we model common aerodynamic effects, including air drag, ground effect, and near-wall effect, and incorporate actuator saturation. 
Real-world experiments further validate selected modeled effects and demonstrate the learning pipeline on hardware.

In summary, our contributions are: (i) modular AM simulation infrastructure; (ii) a reusable benchmark for controlled studies of individual system components and their interactions; and (iii) empirical investigations of high-level policies, policy--control interfaces, and embodiment-dependent behavior, supported by real-world validation. These studies are not intended to provide exhaustive, generalized conclusions across all AM scenarios, but rather to show how modular simulation can isolate design factors and guide future improvements in embodiments, controllers, and policies.

\vspace{-0.1in}

\section{Related Work}
\label{sec: Related Works}

\subsection{Aerial Manipulation}

Reliable AM requires tight integration of hardware design, low-level control, whole-body planning, and high-level decision-making. Unlike their ground-based counterparts, aerial manipulators must compensate for gravity and contact wrenches solely through flight dynamics, motivating various adaptive and robust control strategies to address these challenges~\citep{lee2021aerial, liang2022adaptive, malczyk2023multi, chen2026contact}.
Another line of work focuses on motion planning and hardware design to mitigate the limitations of conventional underactuated multirotors. In these systems, horizontal acceleration is coupled to attitude, so lateral motion during manipulation can perturb end-effector orientation. Whole-body planning and control methods address this issue by coordinating the base and manipulator trajectories~\citep{deng2025whole, marti2021full, lee2020aerial}. Alternatively, fully actuated and overactuated multirotors decouple translation and rotation, simplifying contact-rich manipulation~\citep{bodie2020active, ryll20196d, lee2025autonomous, zhan2026aerial}.
Recently, there has been increasing interest in data-driven control and learning-based strategies, including reinforcement learning and imitation learning~\citep{zhang2022learning, cuniato2023learning, zeng2025learning, deshmukh2025global, tucker2026pimakeflyphysicsguided}. 
These methods are relevant for vision-guided high-level decision-making where manually designing task-specific policies can be difficult. Because data collection is expensive, failure-prone, and difficult to scale, researchers have explored end-effector-centric representations and teleoperation-based data collection to train visuomotor policies~\cite{he2025flying, gupta2025umi}. 
Despite these advances, AM systems are often evaluated on task-specific setups. As a result, it remains difficult to isolate the effects of controller design, robot morphology, and learned policies under shared task settings and disturbance conditions, motivating a standardized benchmark.

\subsection{Robot Learning Benchmarks}

Benchmarks have been instrumental in advancing robot learning by enabling standardized and reproducible evaluation across paradigms such as imitation learning \cite{mandlekar2021matters}, reinforcement learning \cite{james2020rlbench, chen2022towards}, and meta-reinforcement learning \cite{yu2020meta, mclean2025meta}. 
Several benchmarks have expanded task complexity and diversity by scaling to larger task suites \cite{mu2021maniskill, gu2023maniskill2}, long-horizon and sequential tasks \cite{liu2023libero, han2025robocerebra}, and richer sensing modalities such as language and vision \cite{mees2022calvin}. 
Other benchmarks improve real-world relevance and move toward mobile settings through household-scale activity graphs~\cite{li2024behavior}, bimanual manipulation in home-like environments~\citep{chernyadev2024bigym}, and a rich kitchen-manipulation task suite~\citep{nasiriany2026robocasa365largescalesimulationframework}.
However, these benchmarks primarily evaluate high-level policies in relatively stable tabletop or mobile manipulation scenarios. They are not designed to capture how robot embodiment, low-level control, and dynamic disturbances jointly affect task performance.

In dynamics-critical settings such as legged and aerial robotics, benchmarks increasingly evaluate perception and control in addition to policy learning. Recent humanoid benchmarks emphasize whole-body coordination and loco-manipulation~\citep{sferrazza2024humanoidbench}. For aerial robots, navigation-centered benchmarks focus on egocentric vision, open-world goals, and safety-critical control~\citep{gao2025openfly, xiao2025uav, yu2025flightbench}.
\citet{suarez2020benchmarks} presented an AM benchmark focused primarily on the physical capabilities of hardware prototypes, but that work does not include simulation-based evaluation of learned policies. 
AIR-VLA~\citep{sun2026airvlavisionlanguageactionsystemsaerial} recently introduced an aerial manipulation VLA benchmark for evaluating high-level reasoning and task execution. In contrast, \ourbenchmark focuses on system-level evaluation of how control, embodiment, and learning algorithms jointly affect AM performance, and provides real-world validation of simulated behaviors and a hardware test of the learning pipeline. 

\begin{figure*}[ht]
    \centering
    \includegraphics[width=0.99\linewidth, trim={0 45cm 0 0}, clip]{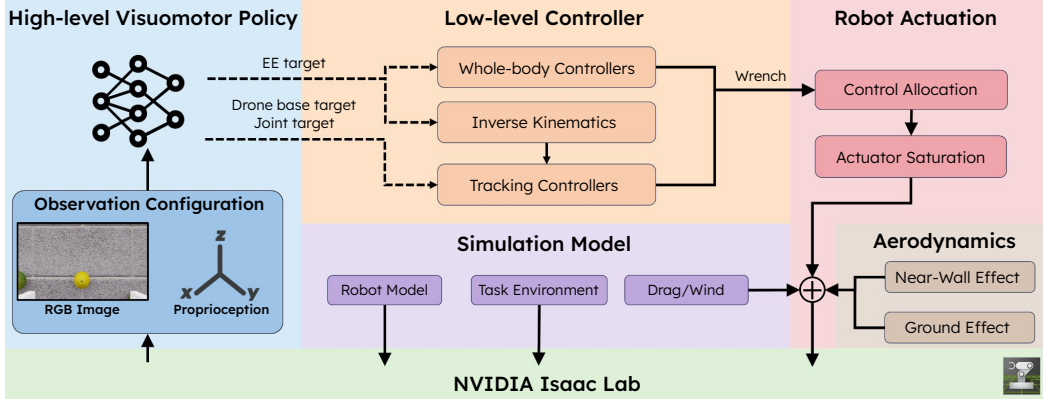}
    \caption{\textbf{System diagram of AM-Bench}. The framework follows a modular pipeline comprising diverse task environments, robot models, disturbances, actuator constraints, low-level controllers, and high-level visuomotor policies within a high-fidelity simulation environment.
    \label{fig: diagram}
    }
    \vspace{-0.0in}
\end{figure*}

\section{Simulation Suite}
\label{sec: Benchmark Suite}

\newlength{\taskimgwidth}
\setlength{\taskimgwidth}{0.16\textwidth}

\newcommand{\TwoStackImg}[2]{%
  \begin{minipage}[t]{\taskimgwidth}
    \centering
    \includegraphics[width=\linewidth]{#1}\vspace{2pt}
    \includegraphics[width=\linewidth]{#2}
  \end{minipage}%
}

\begin{figure*}[t!]
  \centering

  \begin{minipage}[t]{\textwidth}
    \centering
    \TwoStackImg{figures/master_shots_cropped/PressButton_1}
                {figures/master_shots_cropped/PressButton_2}
    \TwoStackImg{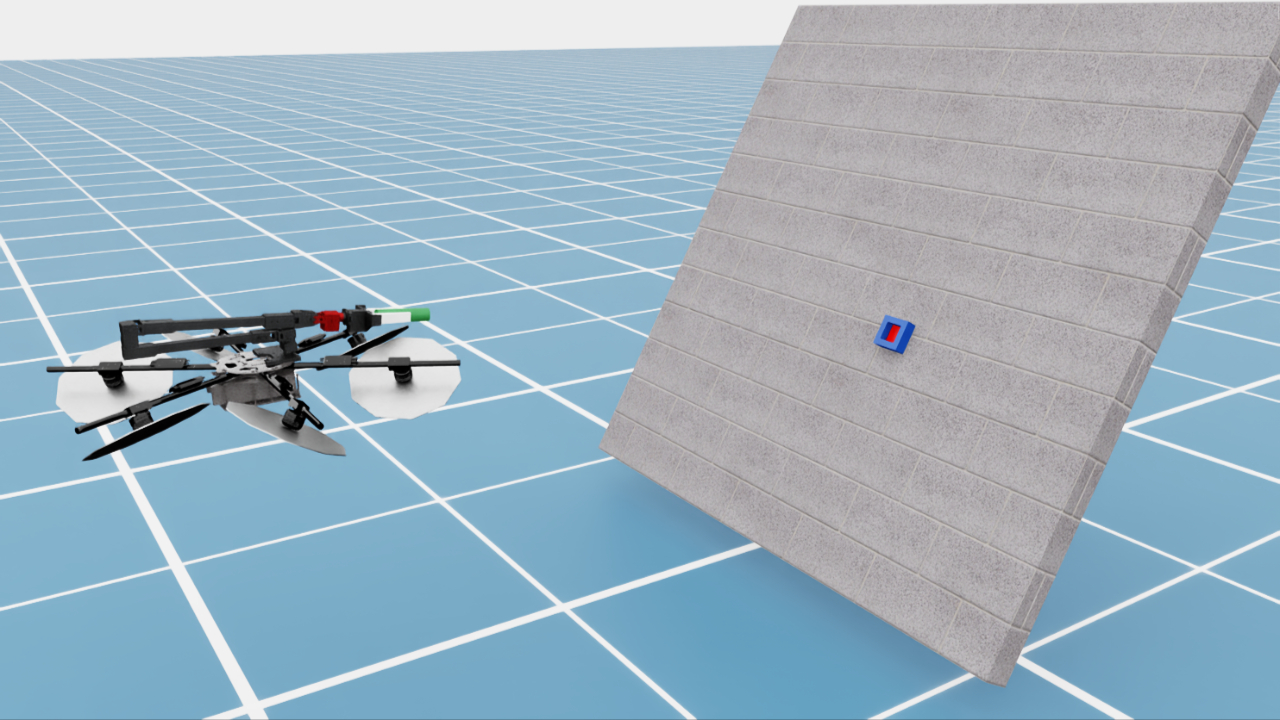}
                {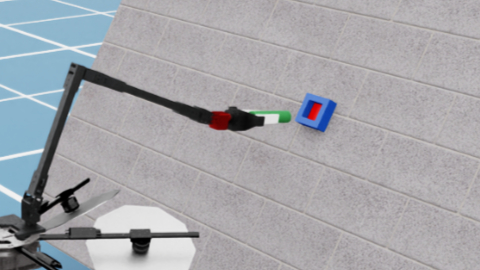}
    \TwoStackImg{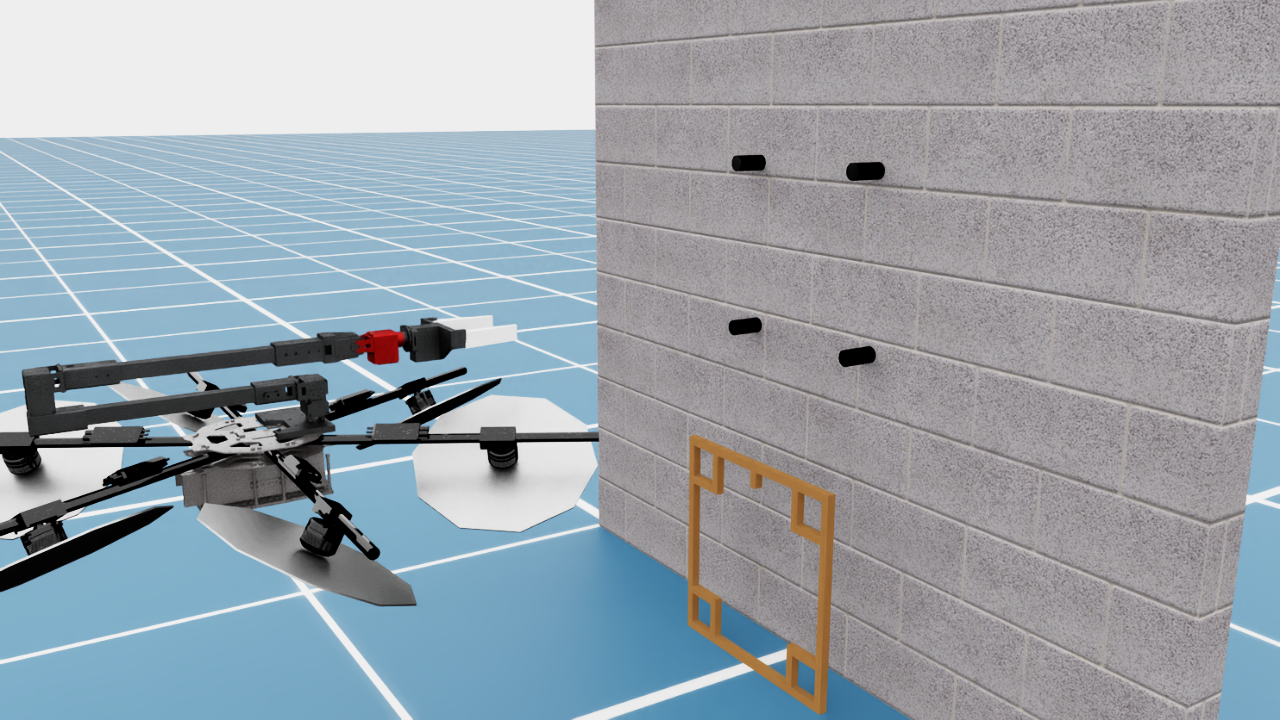}
                {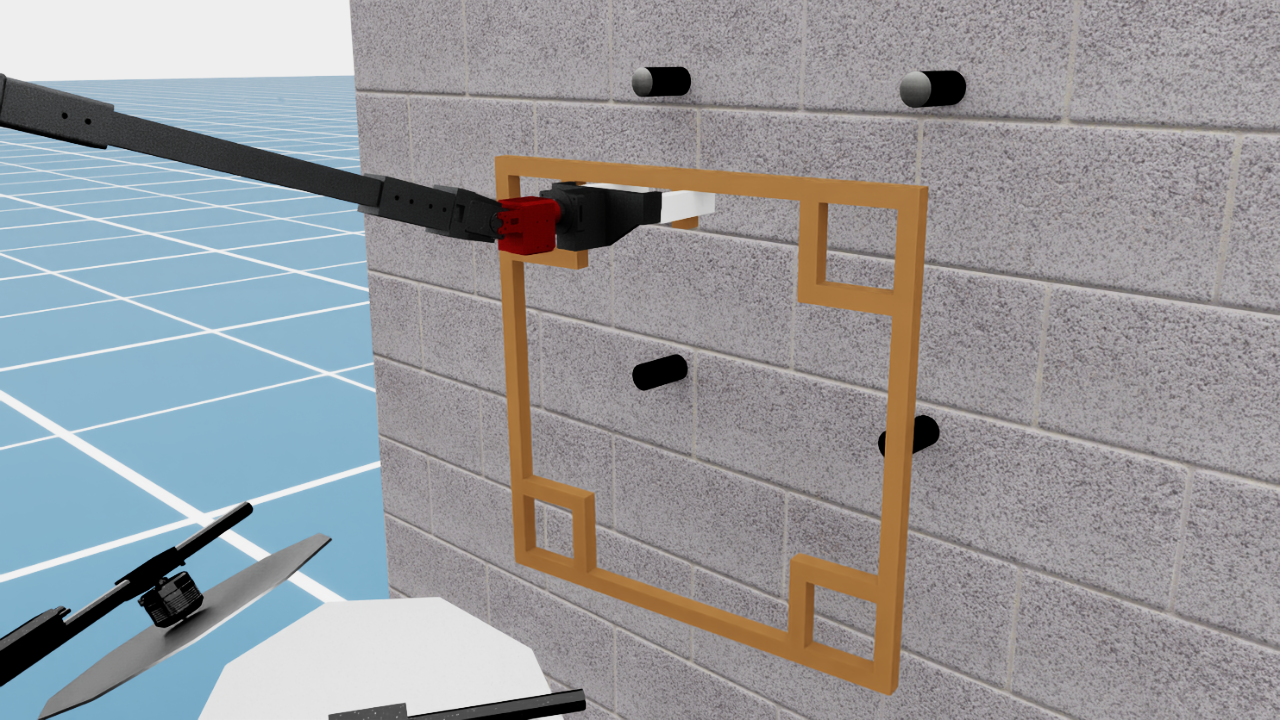}
    \TwoStackImg{figures/master_shots_cropped/CabinetPickPlace_1}
                {figures/master_shots_cropped/CabinetPickPlace_2}
    \TwoStackImg{figures/master_shots_cropped/LemonPickPlace_1}
                {figures/master_shots_cropped/LemonPickPlace_2}
    \TwoStackImg{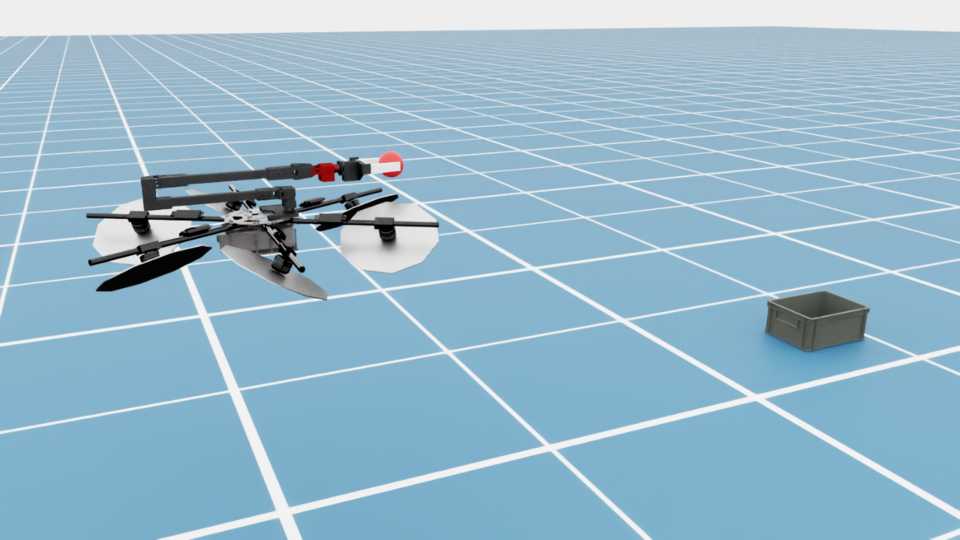}
                {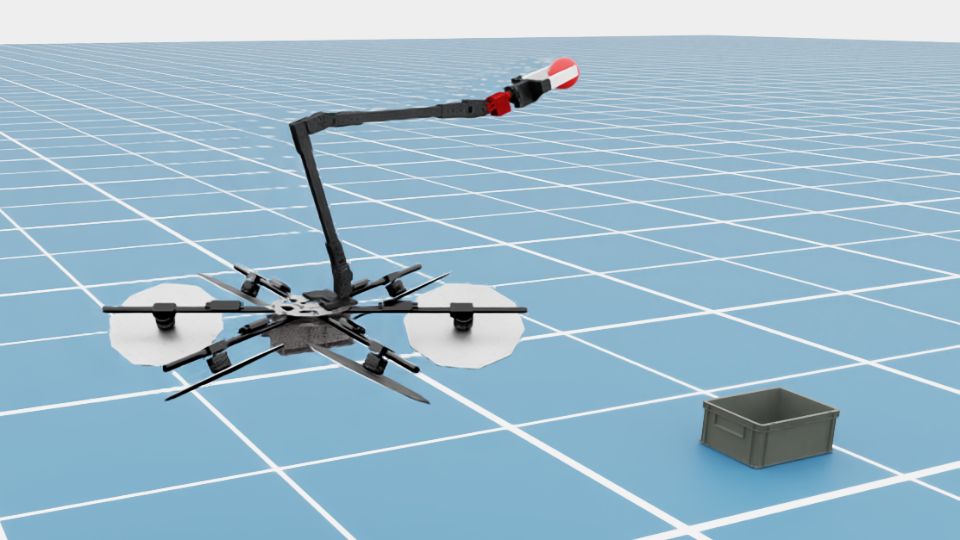}
  \end{minipage}

  \vspace{6pt}

  \begin{minipage}[t]{\textwidth}
    \centering
    \TwoStackImg{figures/master_shots_cropped/RotateValve_1}
                {figures/master_shots_cropped/RotateValve_2}
    \TwoStackImg{figures/master_shots_cropped/PushSlider_1}
                {figures/master_shots_cropped/PushSlider_2}
    \TwoStackImg{figures/master_shots_cropped/PullLever_1}
                {figures/master_shots_cropped/PullLever_2}
    \TwoStackImg{figures/master_shots_cropped/OpenDoor_1}
                {figures/master_shots_cropped/OpenDoor_2}
    \TwoStackImg{figures/master_shots_cropped/WipeSurface_1}
                {figures/master_shots_cropped/WipeSurface_2}
    \TwoStackImg{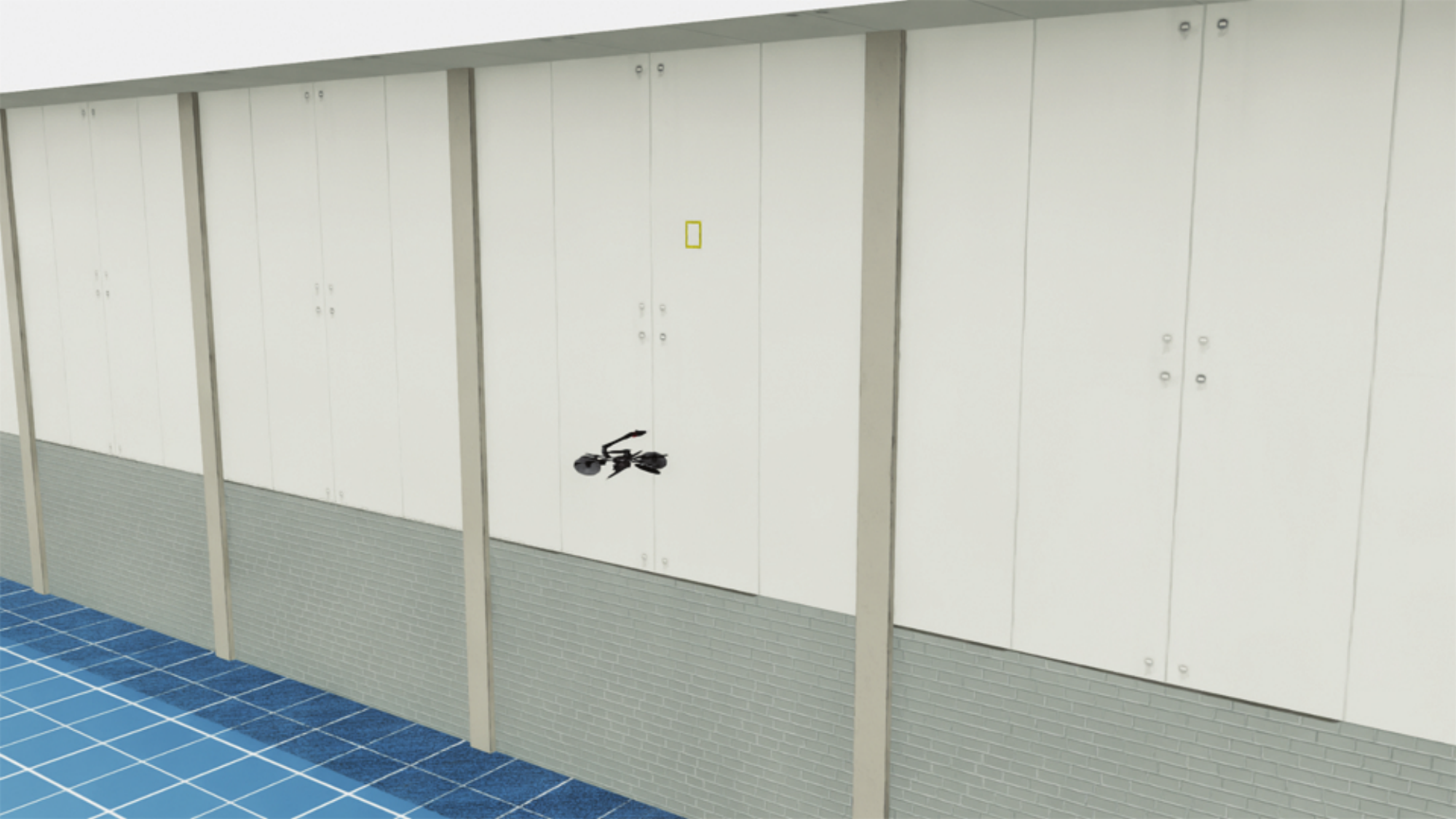}
                {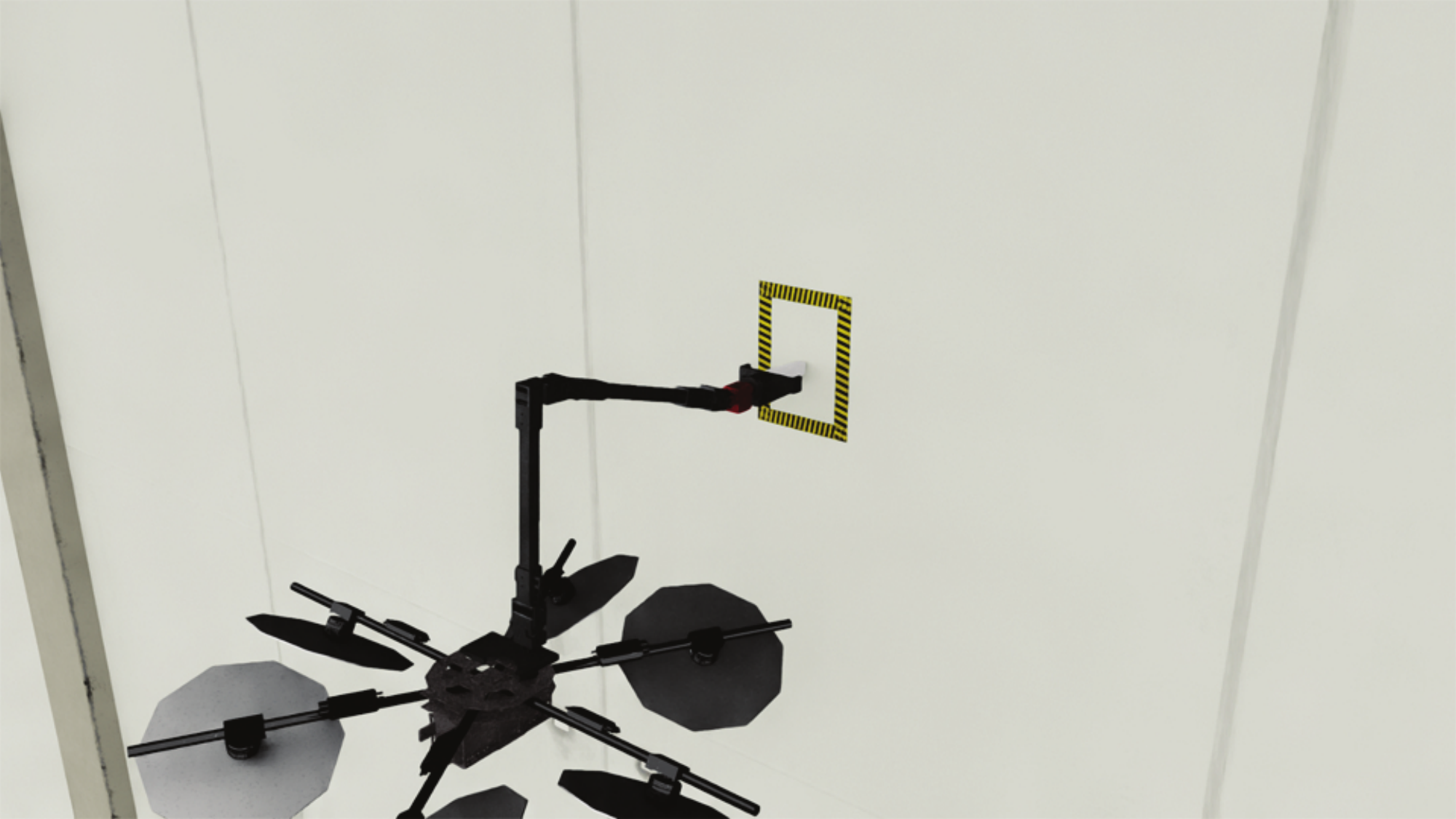}
  \end{minipage}

  \caption{Twelve representative tasks. Top row: \textit{press button}, \textit{peg-in-hole}, \textit{frame assembly}, \textit{cabinet pick-and-place}, \textit{lemon harvesting}, and \textit{toss ball}. Bottom row: \textit{rotate valve}, \textit{push slider}, \textit{pull lever}, \textit{open door}, \textit{wipe window}, and \textit{NDT}.}
  \vspace{-0.2in}
  \label{fig:benchmark-task-groups}
\end{figure*}

We design \ourbenchmark as a modular simulation suite for controlled, system-level evaluation of multirotor-based aerial manipulation. Rather than treating policy learning as an isolated problem, the framework factorizes an AM system into five configuration modules: task environment (Section~\ref{subsec: task-env}), robot platform (Section~\ref{subsec: Platform}), high-level policy interface (Section~\ref{subsec:high-level-interface}), low-level control (Section~\ref{subsec: low-level-control}), and physical disturbances and actuator constraints (Section~\ref{subsec: disturbance}). Fig.~\ref{fig: diagram} illustrates their relationships.

\subsection{Task Environment}
\label{subsec: task-env}

Real-world AM applications often target elevated or hard-to-reach inspection, maintenance, installation, and harvesting scenarios where human access is costly or risky. We organize the suite's 12 representative tasks into three primitive skill categories, as illustrated in Fig.~\ref{fig:benchmark-task-groups}: 

\begin{itemize}[leftmargin=*]
    \item \textbf{Instantaneous interaction}, including \textit{press button} and \textit{peg-in-hole}. These tasks evaluate the robot's ability to complete brief, precise interactions.
    \item \textbf{Object transport}, including \textit{frame assembly}, \textit{cabinet pick-and-place}, \textit{lemon harvesting}, and \textit{toss ball}. These tasks test whether the policy can visually ground the target object and adapt to payload-induced changes in dynamics. \textit{Toss ball} is a dynamic task that requires fast whole-body motion.
    \item \textbf{Articulated object and constrained contact}, including \textit{rotate valve}, \textit{push slider}, \textit{pull lever}, \textit{open door}, \textit{wipe window}, and \textit{NDT}. The aerial robot's motion in $SE(3)$ is constrained by an articulation mechanism or restricted to a low-dimensional manifold. These tasks challenge the policy's ability to comply with imposed trajectories while compensating for contact wrenches that can destabilize the floating base.
\end{itemize}

Built-in domain randomization of variables such as wall pitch, asset texture, and friction enables easy adjustment of task difficulty. 

\subsection{Robot Platform}
\label{subsec: Platform}

\begin{table}[t]
    \centering
    \caption{Supported aerial manipulation platforms in \ourbenchmark.}
    \label{tab: platforms}
    \resizebox{\linewidth}{!}{
    \begin{tabular}{lllclc}
        \toprule
        Category & Platform & Base (\# of actuators) & Arm DoF & Description & Ref. \\ %
        \midrule
        \multirow{2}{*}[-1ex]{UA}
        & \textbf{UA-Quad}
        & \makecell[l]{Quadrotor (4)}
        & 4
        & \makecell[l]{Compact base, lightweight below-base arm; \\ baseline small-payload platform.}
        & \citep{wang2023millimeter} \\
        
        & \textbf{UA-Hexa} 
        & \makecell[l]{Hexarotor (6)}
        & 4
        & \makecell[l]{Larger base, long-reach high-torque above-base arm; \\ alternative underactuated configuration.}
        & 
        \makecell[c]{\citep{he2025flying} \\ (Derived)}  \\
        \midrule
        
        \multirow{2}{*}[-1ex]{FA/OA}
        & \textbf{FA-Hexa} 
        & \makecell[l]{Fixed-tilt hexarotor (6)}
        & 4
        & \makecell[l]{6-DoF wrench control decouples translation and rotation; \\ tests full-actuation benefits for policies using end-effector target commands.}
        & \citep{he2025flying} \\
        
        & \textbf{Omni-Hexa} 
        & \makecell[l]{Variable-tilt multirotor (12)} 
        & 3
        & \makecell[l]{Capable of hovering at an arbitrary orientation; \\ tests enlarged feasible wrench space under rotor saturation.}
        & \citep{lee2025autonomous} \\
        \bottomrule
    \vspace{0.05em}
    \end{tabular}
    }
    {\scriptsize UA: underactuated; FA: fully actuated; OA: overactuated}
        \vspace{-2.0em}

\end{table}

Unlike tabletop manipulation, humanoid, or quadrupeds, the AM field lacks a widely adopted hardware platform, and physically plausible AM systems must satisfy component-level constraints such as payload capacity, manipulator mass, flight time, and actuator limits. Rather than introducing artificial platforms, we prioritize configurations that have been demonstrated or are physically motivated by real-world AM systems. Specifically, we select platforms that comprise a multirotor floating base and a serial manipulator and span the principal multirotor actuation classes: underactuated, fully actuated, and overactuated systems. 
As shown in Fig.~\ref{fig:teaser}, \ourbenchmark supports four multirotor-based AM platforms: \textbf{UA-Quad}, \textbf{UA-Hexa}, \textbf{FA-Hexa}, and \textbf{Omni-Hexa}. 
These include conventional quadrotor- and hexarotor-based underactuated systems, a fixed-tilt fully actuated hexarotor, and an omnidirectional variable-tilt multirotor. 
Brief descriptions are provided in Table~\ref{tab: platforms}, with additional details in Appendix~\ref{app:robots}.
In addition to these four complete AM platforms, we include an \textbf{EE-only} floating end-effector model, similar in spirit to the Universal Manipulation Interface~\cite{chi2024universal}, that serves as an oracle baseline for task execution under the common end-effector target interface.

\subsection{High-Level Policy}
\label{subsec:high-level-interface}

\ourbenchmark supports both IL and VLA paradigms for high-level policies. It exposes a modular interface with configurable RGB and proprioceptive observation spaces, including end-effector and base camera views, end-effector and base states, gripper state, and arm joint angles. Policies can use one of two action interfaces: an \textbf{end-effector target (EE target)} command or a \textbf{base and arm joint target} command. These interfaces enable imitation-learning and VLA policies to be evaluated under shared task, embodiment, and controller settings. Additional interface details are provided in Appendix~\ref{app:policy-interface}.

\subsection{Low-Level Control}
\label{subsec: low-level-control}

Because AM payload constraints often preclude the use of heavy torque-controlled robotic arms, we follow a common approach in the field \cite{lee2021aerial, liang2022adaptive, he2025flying} and assume that the manipulator is equipped with an independent position controller. 
For the multirotor base, we implement wrench-based control. For UA platforms, this produces a control input comprising total thrust and three-axis torque; for FA/OA platforms, it produces a 6-DoF wrench.

Given the high-level action interfaces in Sec.~\ref{subsec:high-level-interface}, we provide two mappings for EE target commands. The \textbf{whole-body MPC} mapping~\cite{he2025flying} jointly computes the required base wrench and desired joint angles in a unified optimization framework, enabling coordinated control under explicit constraints. The \textbf{decoupled IK-tracking} mapping converts an EE target into a desired base pose and arm joint angles; the resulting base trajectory is tracked by either a geometric PID controller~\cite{goodarzi2013geometric} or an $L_1$ adaptive controller~\cite{wu2025L1quad} for disturbance and model-mismatch compensation. For base and arm joint target commands, the specified joint angles are tracked by the manipulator's independent position controller, while the specified base trajectory is tracked by PID or $L_1$ control. Implementation details are provided in Appendix~\ref{app:low-level-control}.

\subsection{Disturbances, Actuator Dynamics, and Saturation}
\label{subsec: disturbance}

Accurately modeling aerodynamic effects near environmental structures is critical when simulating multirotor-based systems. 
Proximity-induced effects, such as ground effect \citep{sanchez2017characterization} and near-wall effect \citep{ding2022passive, zhang2026physics}, can degrade control performance, especially under actuator saturation where commanded rotor thrusts exceed feasible limits. 
To capture this interaction efficiently, \ourbenchmark incorporates reduced-order aerodynamic disturbance models coupled with explicit actuator saturation rather than computationally expensive computational fluid dynamics (CFD). An optional reduced-order actuator module additionally maps each saturated rotor-thrust command to normalized rotor speed, applies a first-order response and a symmetric normalized-speed rate limit, and then maps the realized state back to thrust.

Both ground and near-wall effects are modeled as rotor-level perturbations to the effective thrust. 
The ground-effect model follows \citep{kan2019analysis} and augments each rotor thrust according to the distance to the nearest downward surface along the rotor thrust axis. The near-wall model follows \citep{ding2023aerodynamic} and captures increased vertical thrust and attractive lateral forces near vertical surfaces. 

To incorporate these effects, the control allocation layer first maps a desired body wrench to individual rotor-thrust commands. Physical thrust limits and any configured actuator transients are applied before the rotor-level aerodynamic perturbations. Ground effect is applied first, followed by near-wall effect; the corrected rotor forces and reaction torques are then reconstructed as a body wrench. Finally, body-frame aerodynamic drag \citep{jacinto2023pegasus} and a constant world-frame wind force are added. This signal flow is summarized in Fig.~\ref{fig: diagram}, together with built-in sensor-noise and actuator-command-noise models. The detailed formulations, control-allocation model, and actuator validation are provided in Appendices~\ref{app:disturbance-actuator}, \ref{app:control-allocation}, and~\ref{app:actuator-dynamics-experiment}, respectively.

\vspace{-0.1in}

\section{Experiments}
\label{sec: Experiments}
\vspace{-0.1in}
To further understand the current challenges in AM, we use our simulation suite to investigate the following research questions:
\begin{itemize}[leftmargin=*]
    \item How well do current visuomotor policies, including VLAs, solve representative aerial manipulation tasks under a common execution interface?
    \item How does the choice of policy--control interface affect task performance?
    \item How does aerial-manipulator embodiment affect task execution?
    \item Can \ourbenchmark capture real-world system behavior and support deployment of its learning pipeline on hardware?
\end{itemize}

\subsection{High-Level Visuomotor Policy}
\label{subsec: high-level-policy}

We evaluate both specialist imitation-learning baselines and pretrained VLAs to understand how high-level visuomotor policies perform on AM systems. All methods are evaluated on FA-Hexa using the EE target interface with the IK-PID controller. For each of the $12$ tasks, we collect $80$ successful scripted demonstrations in simulation and aggregate the single-task datasets into a multi-task dataset. We select ACT \cite{zhao2023learning} and DP \cite{chi2023diffusionpolicy} as specialist single-task baselines. For the VLA baselines, we evaluate $\pi_0$~\cite{black2026pi0visionlanguageactionflowmodel} and $\pi_{0.5}$~\cite{intelligence2025pi05visionlanguageactionmodelopenworld} under an adaptation hierarchy: zero-shot evaluation from pretrained checkpoints (ZS), multi-task fine-tuning on the aggregated dataset for embodiment-domain adaptation (MT-FT), and additional task-specific fine-tuning of MT-FT on each single-task dataset (MT $\rightarrow$ ST-FT). Together, these choices span task-specific specialist policies, zero-shot generalists, domain-adapted generalists, and task-specialized generalists. Policy implementation details are provided in Appendix~\ref{app:high-level-policy}.

\begin{wraptable}{r}{0.47\textwidth}
  \vspace{-2.0em}
  \centering
  \caption{High-level policy performance.}
  \label{tab:exp1-high-level}
  \vspace{-0.5em}
  \setlength{\tabcolsep}{4pt}
  \renewcommand{\arraystretch}{0.95}
  \scriptsize
  \begin{tabular}{lcc}
    \toprule
    \textbf{Policy} &
    \textbf{SR \textbf{[\%]} \(\uparrow\)} &
    \textbf{Subtask Completion \textbf{[\%]} \(\uparrow\)} \\
    \midrule
    ACT & 26.94 & 34.17 \\
    DP & 41.39 & 53.52 \\
    \(\pi_0\) ZS & 8.61 & 8.95 \\
    \(\pi_0\) MT-FT & 36.67 & 54.63 \\
    \(\pi_0\) MT $\rightarrow$ ST-FT & 45.00 & 59.04 \\
    \(\pi_{0.5}\) ZS & 2.78 & 6.30 \\
    \(\pi_{0.5}\) MT-FT & 49.17 & 59.44 \\
    \(\pi_{0.5}\) MT $\rightarrow$ ST-FT & 52.22 & 67.31 \\
    \bottomrule
  \end{tabular}
  \vspace{-1.0em}
\end{wraptable}

Each method is evaluated across all 12 tasks with $30$ evaluation rollouts per task, and we report macro-average task success and subtask completion rates in Table~\ref{tab:exp1-high-level}. Direct zero-shot VLA transfer remains limited, indicating a domain gap between the pretrained VLAs and AM embodiments and tasks. This gap is also reflected in the competitiveness of task-specific IL baselines: although $\pi_{0.5}$ achieves the best overall performance after fine-tuning, DP and ACT remain competitive in subtask completion and can outperform VLA variants on some tasks, such as \textit{rotate valve}. Fine-tuning mitigates this gap, with the dominant gains coming from broad AM-domain adaptation: ZS to MT-FT improves macro-average SR by $28.1$ percentage points for $\pi_0$ and $46.4$ points for $\pi_{0.5}$, whereas subsequent task-specific specialization adds $8.3$ and $3.1$ points, respectively. Higher subtask completion rates indicate that the policies learned meaningful intermediate behaviors, such as grasping, but failures remain concentrated in tasks such as \textit{frame assembly}, \textit{rotate valve}, and \textit{toss ball}, suggesting that precise end-effector control and contact-rich interactions remain bottlenecks. Full per-task results are provided in Appendix~\ref{app:high-level-policy-results}.

\subsection{Policy--Control Interface}

\label{subsec:control-policy}

\begin{table}[!ht]
    \vspace{-1em}
  \centering
  \caption{Representative DP low-level interface ablation and mechanism breakdown.}
  \label{tab:exp2_dp_mechanism_main}
  \scriptsize
  \setlength{\tabcolsep}{2.4pt}
  \renewcommand{\arraystretch}{1.12}
  \begin{tabular*}{\textwidth}{@{\extracolsep{\fill}}llccccc@{}}
    \toprule[0.8pt]
    & &
    \multicolumn{1}{c}{\textbf{Outcome}} &
    \multicolumn{2}{c}{\textbf{Tracking}} &
    \multicolumn{2}{c}{\textbf{Limit Use}} \\
    \cmidrule[0.1pt](lr){3-3}
    \cmidrule[0.1pt](lr){4-5}
    \cmidrule[0.1pt](l){6-7}
    \textbf{Task} &
    \textbf{Control Interface} &
    \textbf{SR [\%]  \(\uparrow\)} &
    \makecell{\textbf{Norm. EE} \textbf{Error [$10^{-2}$]} \(\downarrow\)} &
    \makecell{\textbf{Norm. Base} \textbf{Error [$10^{-2}$]} \(\downarrow\)} &
    \makecell{\textbf{Tilt Use}  \textbf{[rad]} \(\downarrow\)} &
    \makecell{\textbf{Sat.} \textbf{Rate [\%]} \(\downarrow\)} \\
    \midrule[0.7pt]
    \multirow{5}{*}[-1ex]{\makecell[l]{Press\\button}}
    & EE target, IK-PID & \textbf{73.3} & $0.250 \pm 0.023$ & $0.139 \pm 0.016$ & $\mathbf{0.010 \pm 0.002}$ & $\mathbf{5.55 \pm 0.78}$\\
    & \makecell[l]{Base and arm joint\\target, PID} & 56.7 & N/A & $\mathbf{0.130 \pm 0.011}$ & $0.094 \pm 0.024$ & $16.59 \pm 1.35$ \\
    \cmidrule[0.1pt]{2-7}
    & EE target, IK-$L_1$ & 83.3 & $0.466 \pm 0.915$ & $\mathbf{0.157 \pm 0.093}$ & $\mathbf{0.047 \pm 0.171}$ & $\mathbf{6.43 \pm 3.86}$ \\
    & \makecell[l]{Base and arm joint\\target, $L_1$} & 83.3 & N/A & $0.256 \pm 0.304$ \ & $0.092 \pm 0.039$ & $20.85 \pm 11.42$ \\
    \cmidrule[0.1pt]{2-7}
    & EE target, MPC & 100.0 & $1.594 \pm 3.476$ & N/A & $0.016 \pm 0.007$ & $1.41 \pm 0.77$\\
    \midrule[0.7pt]
    \multirow{5}{*}[-1ex]{\makecell[l]{Push\\slider}}
    & EE target, IK-PID & \textbf{63.3} & $0.988 \pm 0.316$ & $0.275 \pm 0.087$ & $0.040 \pm 0.012$ & $39.55 \pm 12.10$\\
    & \makecell[l]{Base and arm joint\\target, PID} & 0.0 & N/A & N/A & N/A & N/A \\
    \cmidrule[0.1pt]{2-7}
    & EE target, IK-$L_1$ & \textbf{53.3} & $0.817 \pm 0.412$ & $\mathbf{0.233 \pm 0.089}$ & $\mathbf{0.040 \pm 0.018}$ & $\mathbf{32.34 \pm 11.18}$\\
    & \makecell[l]{Base and arm joint\\target, $L_1$} & 16.7 & N/A & $2.656 \pm 0.602$ & $0.248 \pm 0.044$ & $70.96 \pm 7.34$ \\
    \cmidrule[0.1pt]{2-7}
    & EE target, MPC  & 83.3 & $2.249 \pm 0.182$ & N/A & $0.021 \pm 0.006$ & $3.76 \pm 3.42$\\
    \bottomrule[0.8pt]
  \end{tabular*}
  \vspace{-1em}
\end{table}

Under the hierarchical policy--control structure used in \ourbenchmark, the interface between the policy and controller becomes an important design choice. Here, we compare the two interface choices defined in Section~\ref{subsec:high-level-interface}. In addition to task SR, we analyze tracking error and limit use over successful evaluation rollouts to assess tracking precision and stability margins. Task criteria and action definitions are provided in Appendices~\ref{app:task-environment} and~\ref{app:policy-interface}; controller and metric definitions are provided in Appendices~\ref{app:low-level-control} and~\ref{app:evaluation-metrics}. Full ACT and $\pi_{0.5}$ results are provided in Appendix~\ref{app:policy-control-results}.

As shown in Table~\ref{tab:exp2_dp_mechanism_main}, the EE target interface consistently achieves higher task success rates across different tasks and controller choices. It also generally yields lower tracking error, reduced tilt use, and lower saturation rates. These results suggest that directly exposing all base and arm degrees of freedom to the high-level policy makes policy learning unnecessarily difficult and often leads to inefficient or unstable use of the platform limits. In contrast, the EE target interface provides a more task-relevant abstraction: the policy only needs to reason about the desired manipulation objective in Cartesian space, while the low-level controller handles redundancy resolution, base--arm coordination, and constraint management. This separation produces lower limit use and better task success. EE target control with whole-body MPC outperforms the other evaluated configurations, further suggesting that jointly optimizing robotic-arm and floating-base motions for low-level control is beneficial. Overall, the ablation shows that a high-level policy issuing EE target commands, combined with a capable whole-body low-level controller, provides a more suitable interface for AM than directly commanding base and arm joint targets.

\vspace{-0.1in}
\subsection{Robot Embodiment}
\label{subsec:robot-embodiment}

Task-space control is widely used in AM \cite{he2025flying,deshmukh2025global} because it abstracts embodiment-specific control and allows the policy to focus on the desired end-effector interaction. However, AM embodiments are not dynamically equivalent: identical commands can require different base motions, attitude changes, and control allocation depending on the platform actuation model and arm geometry. The relevant coupled and underactuated dynamics are summarized in Appendices~\ref{app:dynamics-coupling} and~\ref{app:underactuated-dynamics}.

\begin{figure}[t]
  \centering

  \begin{subfigure}[t]{0.49\linewidth}
    \centering
    \includegraphics[width=0.49\linewidth]{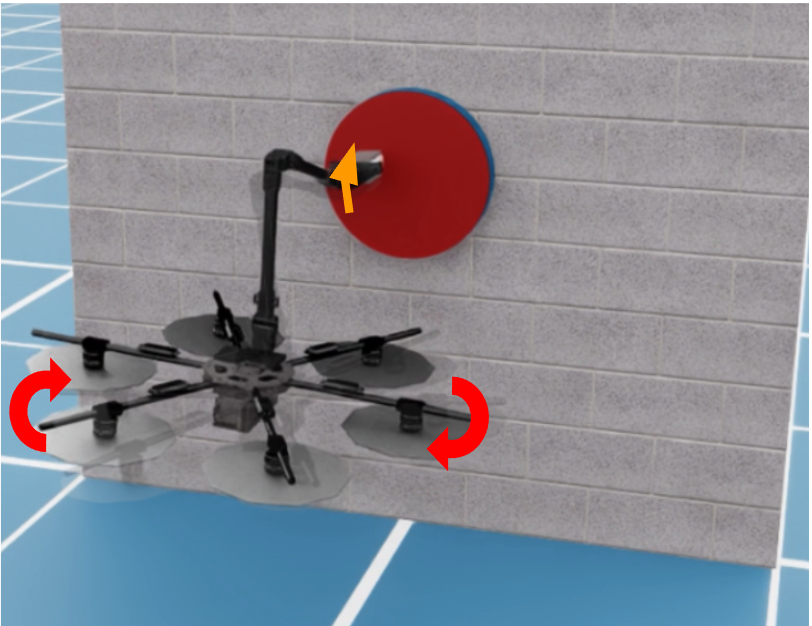}
    \hfill
    \includegraphics[width=0.49\linewidth]{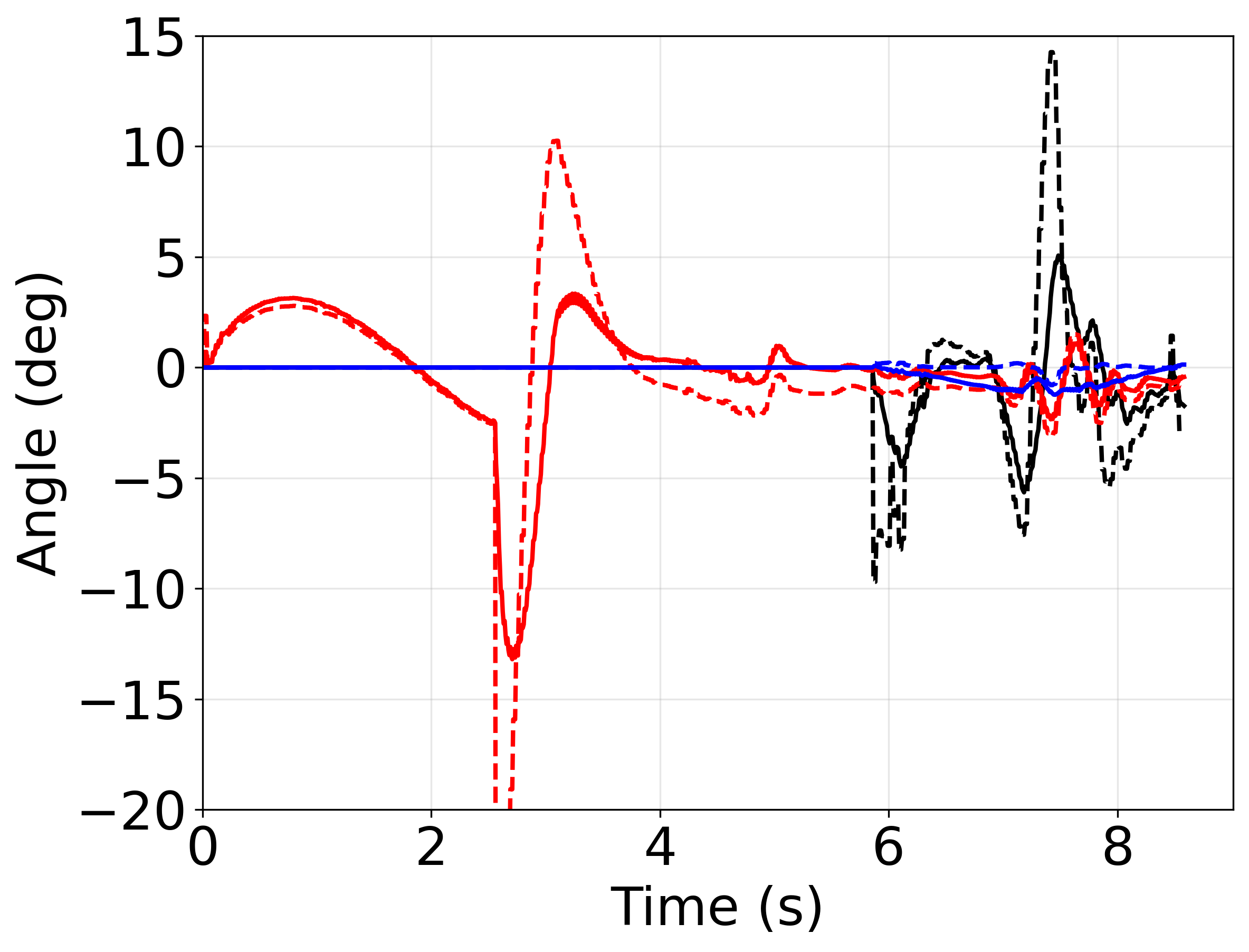}
    \caption{UA-Hexa}
    \label{fig:exp5_ua}
  \end{subfigure}
  \hfill
  \begin{subfigure}[t]{0.49\linewidth}
    \centering
    \includegraphics[width=0.49\linewidth]{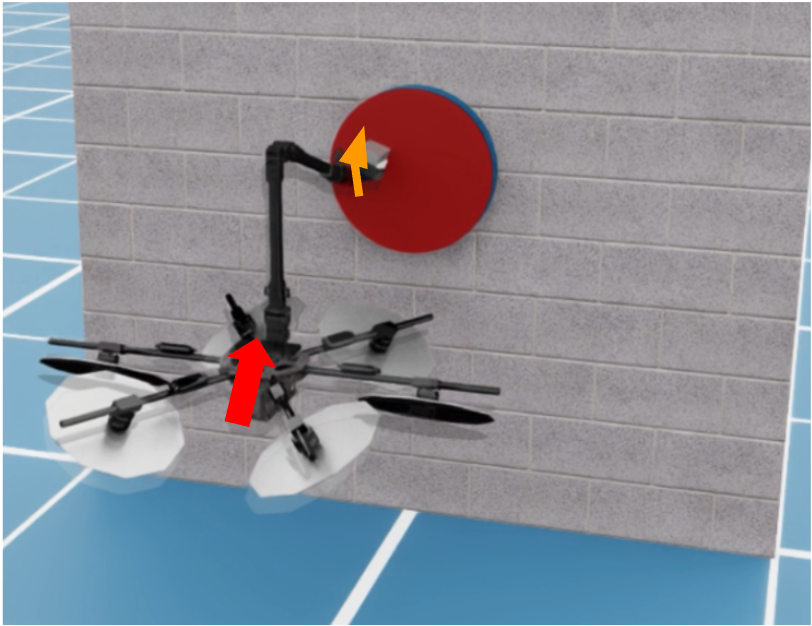}
    \hfill
    \includegraphics[width=0.49\linewidth]{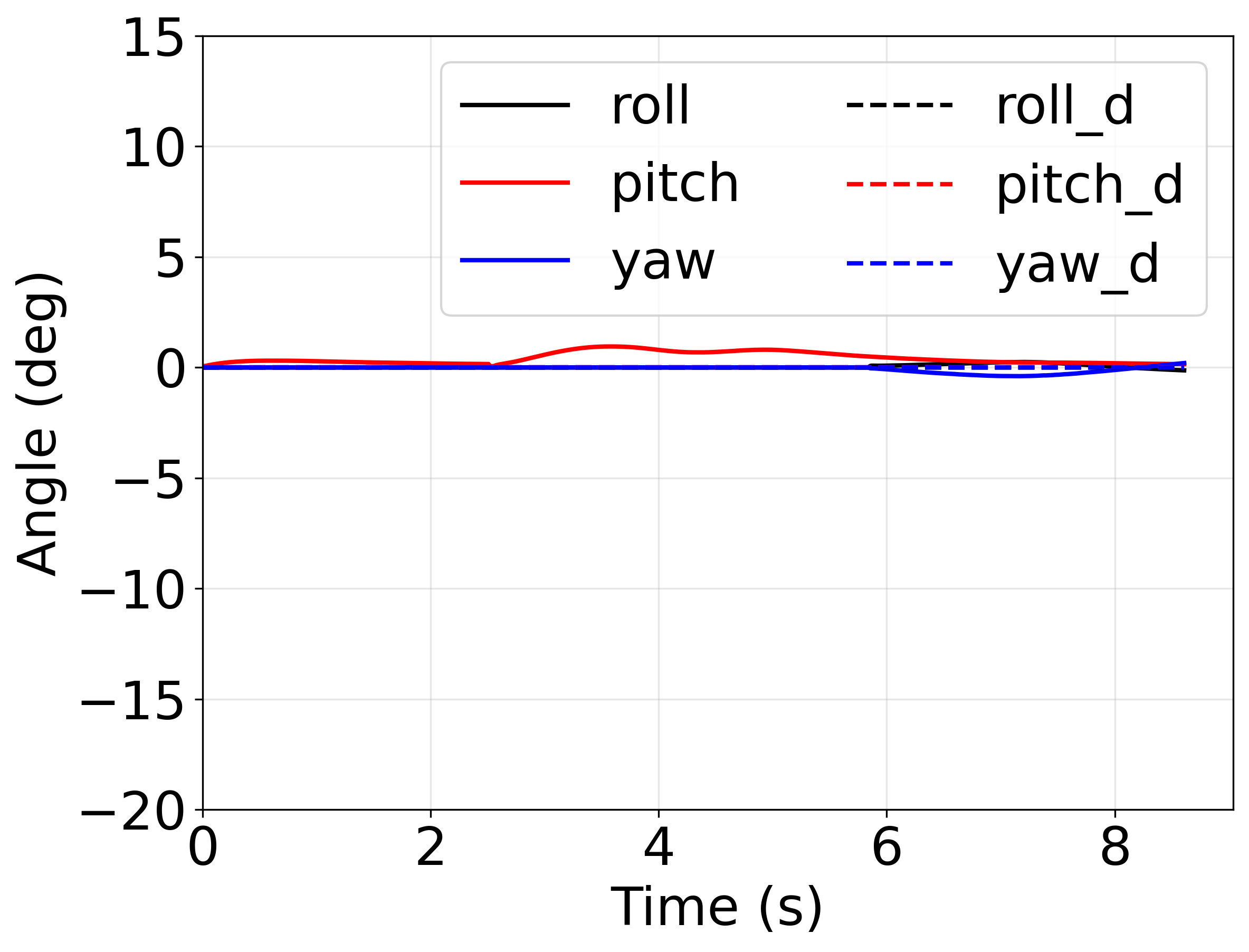}
    \caption{FA-Hexa}
    \label{fig:exp5_fa}
  \end{subfigure}
  \vspace{-0.1in}
  \caption{Visualization of the impact of underactuation during the \textit{rotate valve} task. The time-lapse composite images show configuration changes at $t \approx 7$ seconds in simulation. The plots show the desired and actual orientation trajectories of the drone base during the task.}
  \label{fig:exp5}
  \vspace{-0.2in}
\end{figure}

\newpage
\begin{wraptable}[10]{r}{0.48\textwidth}
  \centering
  \captionsetup{skip=2pt}
  \caption{Embodiment comparison for selected tasks.}
  \label{tab:exp3}
  \scriptsize
  \setlength{\tabcolsep}{2.0pt}
  \renewcommand{\arraystretch}{0.92}
  \begin{tabular}{llcccc}
    \toprule
    \textbf{Task} &
    \textbf{Metric} &
    \textbf{UA-Q} &
    \textbf{UA-H} &
    \textbf{FA-H} &
    \textbf{Omni-H} \\
    \midrule
    \multirow{4}{*}{\shortstack[c]{Push\\slider}}
    & Norm. EE err. [$10^{-2}$] \(\downarrow\) & 7.506 & 7.051 & \textbf{0.990} & 3.502 \\
    & Norm. base err. [$10^{-2}$] \(\downarrow\) & 4.668 & 7.828 & \textbf{0.275} & 3.223 \\
    & Tilt use [rad] \(\downarrow\) & 0.187 & 0.143 & \textbf{0.057} & 1.160 \\
    & Sat. rate [\%] \(\downarrow\) & 48.01 & 87.44 & 39.52 & \textbf{9.98} \\
    \midrule
    \multirow{4}{*}{\shortstack[c]{Open\\door}}
    & Norm. EE err. [$10^{-2}$] \(\downarrow\) & 6.643 & 2.061 & \textbf{1.185} & 3.483 \\
    & Norm. base err. [$10^{-2}$] \(\downarrow\) & 3.095 & 1.470 & \textbf{0.738} & 1.334 \\
    & Tilt use [rad] \(\downarrow\) & 0.325 & 0.252 & \textbf{0.028} & 1.722 \\
    & Sat. rate [\%] \(\downarrow\) & 45.84 & 77.63 & 27.64 & \textbf{5.37} \\
    \bottomrule
  \end{tabular}
\end{wraptable}

To examine these embodiment-dependent differences, we evaluate different AM platforms under the same EE target objectives using the same scripted high-level policy and IK-PID tracking controller. As shown in Table~\ref{tab:exp3}, underactuated platforms such as UA-Hexa and UA-Quad require roll and pitch motion to generate lateral translation, resulting in larger tilt angles. In contrast, FA-Hexa can translate without tilting its base. Omni-Hexa shows larger tilt angles because it supports tilted hover, which the IK solver exploits for tracking and constraint satisfaction near the task surface. Its much lower saturation further indicates a larger achievable wrench space.

We further compare FA-Hexa and UA-Hexa with the same base frame, arm design, mass, and inertia, differing only in rotor tilt. In the \textit{rotate valve} task, UA-Hexa relies on base attitude changes to produce lateral end-effector motion, introducing additional end-effector rotation, whereas FA-Hexa avoids this coupling through attitude-independent lateral translation, as shown in Fig.~\ref{fig:exp5}. 

Across these selected cases, the common EE target interface exposes embodiment-dependent differences in tracking, tilt use, and rotor saturation.

\subsection{Real-World Experiments}

\begin{figure}[h]
  \centering

  \begin{minipage}[t]{0.44\linewidth}
    \vspace{0pt}
    \centering
    \includegraphics[height=1.04in]{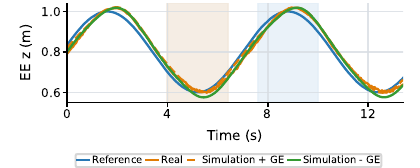}
    \caption{Ground-effect comparison under a shared reference; shaded bands mark the two RMSE intervals.}
    \label{fig:realworld_groundeffect}
  \end{minipage}
  \hfill
  \begin{minipage}[t]{0.54\linewidth}
    \vspace{0pt}
    \centering
    \includegraphics[width=\linewidth]{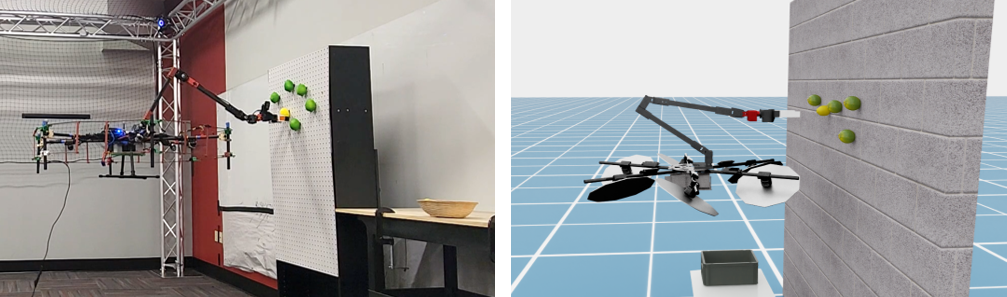}
    \caption{Hardware instantiation of the lemon-harvesting demonstration collection, DP training, and deployment pipeline.}
    \label{fig:real_world_lemonpnp}
  \end{minipage}

  \vspace{-0.1in}
\end{figure}

We conduct real-world experiments to assess whether the simulation captures relevant system behavior and whether its learning pipeline can be instantiated on hardware. For the ground-effect comparison in Fig.~\ref{fig:realworld_groundeffect}, the physical run and both simulated runs follow the same nominal vertical end-effector path through the EE target interface with IK-PID control. The simulated responses are evaluated with and without the ground-effect model. Over the shaded near-ground interval ($z_{\mathrm{ref}}\leq0.7\,\mathrm{m}$), including ground effect reduces the simulation-to-real EE-$z$ RMSE from $1.91$ to $0.99\,\mathrm{cm}$, a $48\%$ reduction. Over the shaded higher-altitude interval ($z_{\mathrm{ref}}\geq0.9\,\mathrm{m}$), the corresponding errors are comparable at $0.75\,\mathrm{cm}$ without and $0.84\,\mathrm{cm}$ with ground effect. Additional details are provided in Appendix~\ref{app:real-world-ground-effect}. For lemon harvesting, we instantiate the demonstration collection, DP training, and EE target deployment pipeline on a physical FA-Hexa, as shown in Fig.~\ref{fig:real_world_lemonpnp}; details are provided in Appendix~\ref{app:real-world-lemon}.

\vspace{-0.1in}

\FloatBarrier

\section{Conclusion} 
\label{sec:conclusion}

We present \ourbenchmark, a modular simulation benchmark for system-level evaluation of policy learning in multirotor-based AM. We integrate task environments, robot embodiments, physical disturbances, low-level controllers, and high-level policy interfaces, enabling controlled evaluation of factors that are difficult to isolate in task-specific AM systems. Our experiments characterize the performance of IL and VLA baselines, the effects of policy--control interfaces, and embodiment-dependent differences in AM, while real-world experiments validate selected modeled effects and instantiate the learning pipeline on hardware. Overall, \ourbenchmark provides a foundation for studying multiple design factors in dynamics-critical manipulation.

\section{Limitations} 
\label{sec:limitations}

While \ourbenchmark provides a modular framework for studying system-level interactions in AM, several limitations remain. 
The benchmark focuses on single-robot, multirotor-based aerial manipulators with rigid arms and does not cover cooperative systems for cable-suspended payload manipulation~\cite{sun2025agile}, continuum or compliant manipulators~\cite{ubellacker2023aggressive}, or soft grippers~\cite{zhao2022modular}. Supporting these morphologies would require different dynamics models and evaluation protocols, which we leave for future work. Our current studies focus on IL and VLA baselines, although the suite includes the infrastructure needed to support reinforcement learning~\cite{deshmukh2025global}. We do not emphasize RL because of the need for careful reward design and the challenges of sim-to-real transfer, although we view it as a promising direction for AM. 
Although the simulator models selected aerodynamic effects and reduced-order actuator dynamics, it does not capture full motor--propeller and ESC dynamics or the complete range of real-world disturbances. Evaluating higher-fidelity models and their effects on learned-policy robustness and task success remains future work, as does comprehensive sim-to-real validation across tasks and platforms.

\clearpage
\acknowledgments{This work was supported in part by the National Science Foundation under Award No. 2541976. Any opinions, findings, and conclusions or recommendations expressed in this material are those of the authors and do not necessarily reflect the views of the National Science Foundation. Dongjae Lee was supported by Basic Science Research Program through the National Research Foundation of Korea (NRF) funded by the Ministry of Education (grant number RS-2025-02634317). Guanya Shi holds concurrent appointments as an Assistant Professor at Carnegie Mellon University and as an Amazon Scholar. This paper describes work performed at Carnegie Mellon University and is not associated with Amazon.}

\bibliography{references}  %

\clearpage
\newpage

\appendix

\section{Benchmark Suite Details}
\label{app:benchmark-suite}

\subsection{Task Environment}
\label{app:task-environment}
Table~\ref{tab:tasks} provides detailed information about each task, including its success criteria, binary subtask criteria, and domain-randomization configuration.

\begin{table}[H]
\centering
\caption{Details of the aerial manipulation task suite.}
\label{tab:tasks}
\scriptsize
\setlength{\tabcolsep}{3.2pt}
\renewcommand{\arraystretch}{1.18}
\begin{tabularx}{\textwidth}{@{}
>{\raggedright\arraybackslash}p{1.9cm}
>{\raggedright\arraybackslash}X
>{\raggedright\arraybackslash}p{3.2cm}
>{\raggedright\arraybackslash}X
@{}}
\toprule
\textbf{Task} & \textbf{Success Criteria} & \textbf{Subtask Completion Criteria} & \textbf{Domain Randomization} \\
\midrule

Press button &
The position of the button's prismatic joint exceeds the pressed threshold. &
\emph{None} &
Wall position and pitch, wall texture, button position, button size, button color. \\

Peg-in-hole &
The peg tip penetrates past the hole surface and remains within the lateral boundaries. &
\emph{None} &
Wall position and pitch, wall texture, hole position, hole size, hole color. \\

Frame assembly &
The frame is aligned with all peg centers and placed onto the pegs. &
Frame lifted; frame placed on pegs. &
Wall position and pitch, wall texture, frame initial position, peg positions. \\

Cabinet pick-and-place &
The can is placed stably on the drawer. &
Door opened; can lifted; can placed on drawer. &
Can position, cabinet door friction, can mass. \\

Lemon harvesting &
The lemon is placed within the container. &
Lemon grasped and detached; lemon delivered and released. &
Wall position, wall texture, lemon and lime positions. \\

Rotate valve &
The angle of the valve-shaft joint reaches the target rotation. &
Valve engaged; valve rotated. &
Wall position and pitch, wall texture, valve position, valve color. \\

Push slider &
The slider's prismatic joint reaches the target position on the rail. &
Slider engaged; slider pushed. &
Wall position and pitch, wall texture, slider position, slider friction, slider color. \\

Pull lever &
The lever's joint angle exceeds the threshold from its initial position. &
Lever engaged; lever pulled. &
Wall position and pitch, wall texture, lever position, lever color. \\

Open door &
The hinge joint angle exceeds the open threshold. &
Door engaged; door opened. &
Door position and orientation, door texture, hinge damping coefficient. \\

Wipe window &
All stains are removed after sufficient contact with force and proximity constraints. &
Individual stains removed. &
Window position, stain positions, window friction. \\

Toss ball &
The ball is tossed into the container while the drone base remains behind the container by the specified distance. &
Ball released; ball delivered. &
Container position. \\

NDT &
The gripper establishes contact with the inspection region and maintains a stable interaction for a fixed period of time. &
\emph{None} &
Location of the inspection point on the building.\\

\bottomrule
\end{tabularx}
\end{table}

\subsection{Robot Platforms}
\label{app:robots}

\begin{table}[H]
\centering
\small
\caption{Physical specifications of the supported aerial manipulation platforms.}
\label{table:specifications}
\begin{tabular}{lcccc}
\hline
\textbf{Parameter} & \textbf{UA-Quad} & \textbf{UA-Hexa} & \textbf{FA-Hexa} & \textbf{Omni-Hexa} \\ \hline
Total Mass [kg] & 4.41 & 6.64 & 6.64 & 	
3.35 \\
MoI ($J_{\mathrm{sys},xx}$) [kg$\cdot$m$^2$] & 0.0885 & 0.212 & 0.212 & 	
0.118  \\
MoI ($J_{\mathrm{sys},yy}$) [kg$\cdot$m$^2$] & 0.123 & 0.252 & 0.252 & 0.121 \\
MoI ($J_{\mathrm{sys},zz}$) [kg$\cdot$m$^2$] & 0.144 & 0.503 & 0.503 & 0.206 \\
Max Thrust (per rotor) [N] & 23.0 & 23.0 & 23.0 & 23.0 \\
Rotor arm length [m] & 0.300 & 0.375 & 0.375 & 0.178 \\
Max manipulator reach [m] & 0.378 & 0.898 & 0.898 & 0.500 \\
Manipulator DoF & 4 & 4 & 4 & 3 \\
\hline
\end{tabular}
\end{table}

We support four aerial manipulator platforms---UA-Quad, UA-Hexa, FA-Hexa, and Omni-Hexa---as illustrated in Fig.~\ref{fig:teaser}. In addition to these complete systems, we provide an EE-only floating end-effector model, similar in spirit to the Universal Manipulation Interface~\cite{chi2024universal}, that serves as an oracle baseline for task execution under the common EE target interface. The detailed physical specifications of the UA-Quad, UA-Hexa, FA-Hexa, and Omni-Hexa platforms are summarized in Table~\ref{table:specifications}. %
The total mass of each platform includes the multirotor base and integrated manipulator arm. The moment of inertia (MoI) of each system is calculated with the manipulator in its zero configuration, in which all joint angles are set to zero. The geometric design of each platform is characterized by its rotor arm length, defined as the distance from the multirotor's geometric center to the center of each rotor, and its maximum manipulator reach, defined as the distance from the first joint to the end-effector. %

\subsection{High-Level Policy Interface Details}
\label{app:policy-interface}

\subsubsection*{Action Space}

Policies can use one of two action interfaces:
\begin{itemize}[leftmargin=*]
    \item \textbf{EE target command}: This task-agnostic, low-dimensional interface is defined in end-effector space, supports intuitive teleoperation, and simplifies visuomotor policy learning~\cite{chi2024universal,he2025flying}. The action specifies a target end-effector pose and gripper command:
    \[
    \mathbf{a}_{\mathrm{ee}}
    =
    [x_{\mathrm{ee}}, y_{\mathrm{ee}}, z_{\mathrm{ee}}, q_{\mathrm{ee},w}, q_{\mathrm{ee},x}, q_{\mathrm{ee},y}, q_{\mathrm{ee},z}, g_{\mathrm{grip}}]^\top
    \in \mathbb{R}^{8},
    \]
    where orientation is represented as a unit quaternion and $g_{\mathrm{grip}}$ is a binary open/close gripper command.
    
    \item \textbf{Base and arm joint target command}: This interface allows the policy to command the drone base pose and manipulator joint angles simultaneously, providing flexibility for whole-body coordination tasks. The target base pose comprises position and quaternion orientation, together with target manipulator joint angles. The action dimension is $8+n_a$, where $n_a$ is the number of manipulator joints:
    \[
    \mathbf{a}_{b\theta}
    =
    [x_b, y_b, z_b, q_{b,w}, q_{b,x}, q_{b,y}, q_{b,z}, g_{\mathrm{grip}}, \theta_{a,1}, \theta_{a,2}, \ldots, \theta_{a,n_a}]^\top
    \in \mathbb{R}^{8+n_a}.
    \]
\end{itemize}

\subsubsection*{Observation Space}
The observation space consists of:
\begin{itemize}[leftmargin=*]
    \item \textbf{Proprioceptive observations}, including end-effector state, base state, gripper state, and arm joint angles.
    \item \textbf{Visual observations} from two RGB cameras: one mounted on the end-effector and the other on the base of each robot platform described in Section~\ref{subsec: Platform}. Camera poses can be adjusted for additional robot platforms.
\end{itemize}

To support straightforward ablations and flexible policy design, the simulation suite provides a modular system for configuring observations. Users can selectively enable a subset of observation components, such as using only arm joint angles for proprioception. 

\subsection{Low-Level Control Implementation Details}
\label{app:low-level-control}

\subsubsection{Inverse Kinematics}

For whole-body motion generation, we formulate a nonlinear least-squares inverse kinematics (IK) problem using the \texttt{Pyroki} toolkit \cite{kim2025pyroki}. At each control step, the IK jointly optimizes the aerial base pose and the manipulator joint angles so that the end-effector tracks a desired pose while maintaining smooth whole-body motion and satisfying kinematic safety constraints. The problem is warm-started from the previous IK solution and solved using a Levenberg--Marquardt solver.

Let the robot have $n_a$ manipulator joints. The IK decision variables are the desired manipulator joint configuration $\boldsymbol{\theta}_{a,d}\in\mathbb{R}^{n_a}$ and the floating
base pose $\mathcal{T}_d\in SE(3)$, where
\begin{equation}
\mathcal{T}_d
=
\left(
\mathbf{R}_d,
\mathbf{p}_d
\right),
\qquad
\mathbf{R}_d \in SO(3),
\quad
\mathbf{p}_d \in\mathbb{R}^{3}.
\end{equation}
Given the desired end-effector pose
$\mathcal{T}_{\mathrm{ee},d} = (\mathbf{R}_{\mathrm{ee},d}, \mathbf{p}_{\mathrm{ee},d})\in SE(3)$, the IK problem is written as
\begin{subequations}
\label{eq:ik_problem}
\begin{align}
\boldsymbol{\theta}_{a,d}^{\star}, \mathcal{T}_d^{\star}
=
\operatorname*{arg\,min}_{\boldsymbol{\theta}_{a,d},\,\mathcal{T}_d}
\quad &
\ell_{\mathrm{ik}}
\\
\mathrm{s.t.}\quad
& \left( \mathcal{T}_d, \boldsymbol{\theta}_{a,d} \right) \in \mathcal{X}_{\mathrm{IK}}
\end{align}
\end{subequations}
Here, the IK objective $\ell_{\mathrm{ik}}$ is implemented as a weighted nonlinear least-squares cost:
\begin{equation}
\begin{aligned}
\ell_{\mathrm{ik}}
=&
\left\|
\mathbf{p}_{\mathrm{ee}}
\left(
\boldsymbol{\theta}_{a,d},
\mathcal{T}_d
\right)
-
\mathbf{p}_{\mathrm{ee},d}
\right\|_{\mathbf{Q}_{p}}^{2}
+
\left\|
\mathbf{e}_R^{\mathrm{IK}}\left( \mathbf{R}_{\mathrm{ee}}(\boldsymbol{\theta}_{a,d},\mathbf{R}_d), \mathbf{R}_{\mathrm{ee},d} \right)
\right\|_{\mathbf{Q}_{R}}^{2}
+
\left\|
\boldsymbol{\theta}_{a,d}
-
\boldsymbol{\theta}_{a,d,\mathrm{prev}}
\right\|_{\mathbf{Q}_{\Delta\theta}}^{2}
+\\
&
\left\|
\mathbf{p}_d-\mathbf{p}_{d,\mathrm{prev}}
\right\|_{\mathbf{Q}_{\Delta p}}^{2}
+
\left\|
\mathbf{e}_R^{\mathrm{IK}}\left(\mathbf{R}_d, \mathbf{R}_{d,\mathrm{prev}}\right)
\right\|_{\mathbf{Q}_{\Delta R}}^{2} 
+
\left\|
\boldsymbol{\theta}_{a,d}
-
\boldsymbol{\theta}_{a,0}
\right\|_{\mathbf{Q}_{\theta}}^{2}
\end{aligned}
\end{equation}
where $\|\mathbf{a}\|_{\mathbf{M}}^{2}=\mathbf{a}^{\top}\mathbf{M}\mathbf{a}$.
The end-effector position $\mathbf{p}_{\mathrm{ee}}$ and orientation $\mathbf{R}_{\mathrm{ee}}$ are computed from the whole-body forward kinematics. The IK orientation error is computed from the relative rotation between two orientations, $\mathbf{R}_1$ and $\mathbf{R}_2$:
\begin{equation}
\mathbf{e}_{R}^{\mathrm{IK}}(\mathbf{R}_1, \mathbf{R}_2)
=
\mathrm{Log}
\left(
\mathbf{R}_{1}^{\top}
\mathbf{R}_{2}
\right),
\end{equation}
where $\mathrm{Log}:SO(3)\rightarrow\mathbb{R}^{3}$ denotes the logarithm map on
$SO(3)$. The smoothness terms penalize deviation from the previous IK solution
$\left(\boldsymbol{\theta}_{a,d,\mathrm{prev}}, \mathcal{T}_{d,\mathrm{prev}} \right)$. The nominal arm posture $\boldsymbol{\theta}_{a,0}$ is used to resolve redundancy and keep the manipulator near a well-conditioned configuration. 

$\mathcal{X}_{\mathrm{IK}}$ bounds the base position, base roll and pitch angles, and the manipulator joint angles. The task-dependent base position bounds keep the aerial platform within a safe workspace. In particular, the bound along the interaction direction is placed with a safety margin from the target object, such as a wall, door, cabinet, or window, while the vertical bound keeps the base above the ground.

\subsubsection{Geometric PID Base Tracking}
\label{app:pid-l1-control}

Both the IK-based EE target interface and the direct base and arm joint target interface provide a desired base pose
$(\mathbf p_d,\mathbf R_d)$ and desired manipulator joint positions
$\boldsymbol\theta_{a,d}$. The base controller receives the measured position
$\mathbf p$, orientation $\mathbf R$, world-frame linear velocity $\mathbf v$, and
body-frame angular velocity $\boldsymbol\omega$, and returns a desired body wrench.
For the fully actuated platforms, we define
\begin{equation}
\begin{gathered}
\mathbf e_p=\mathbf p_d-\mathbf p, \qquad
\mathbf e_v=\mathbf v_d-\mathbf v, \\
\mathbf e_R^{\mathrm{PID}}=\frac{1}{2}\left(\mathbf R_d^\top\mathbf R-\mathbf R^\top\mathbf R_d\right)^\vee,
\qquad
\mathbf e_\omega=\boldsymbol\omega-\mathbf R^\top\mathbf R_d\boldsymbol\omega_d .
\end{gathered}
\end{equation}
The integral states are accumulated from $\mathbf e_p$ and $\mathbf e_R^{\mathrm{PID}}$. The
implemented geometric PID law is
\begin{align}
\mathbf F^w
&=m_{\mathrm{sys}}\left(\mathbf a_d+\mathbf K_p\mathbf e_p+\mathbf K_v\mathbf e_v
+\mathbf K_i\!\int\!\mathbf e_p\,dt+\mathbf g^w\right), \\
\boldsymbol\tau^b
&=\mathbf J_{\mathrm{sys}}\left(-\mathbf K_R\mathbf e_R^{\mathrm{PID}}-\mathbf K_\omega\mathbf e_\omega
-\mathbf K_{I,R}\!\int\!\mathbf e_R^{\mathrm{PID}}\,dt\right)
+\boldsymbol\omega\times\mathbf J_{\mathrm{sys}}\boldsymbol\omega+\boldsymbol\tau_{\mathrm{ff}},
\end{align}
where $m_{\mathrm{sys}}$ and $\mathbf J_{\mathrm{sys}}$ are the aggregate system mass and inertia, and $\mathbf g^w$ is the positive world-frame gravity-compensation vector. The term
$\boldsymbol\tau_{\mathrm{ff}}$ accounts for the desired angular velocity and
acceleration. The commanded body force is $\mathbf F^b=\mathbf R^\top\mathbf F^w$.
Desired velocities and accelerations are obtained by finite differences of the pose
targets.

The underactuated platforms use the same position and attitude error structure in a
nested 4-DoF controller. The position loop constructs a desired thrust direction and
heading, but the commanded body force is restricted to
$\mathbf F^b=[0,0,(\mathbf F^w)^\top\mathbf R\mathbf e_3]^\top$.

\subsubsection{$L_1$ Adaptive Base Tracking}

The $L_1$ controller uses the same state and command interfaces as the 6-DoF PID
controller. Its nominal controller retains the proportional, derivative, gravity,
Coriolis, and angular feedforward terms above, but sets both integral gains to zero.
It augments this nominal wrench with adaptive force and torque estimates, following
the state-predictor, piecewise-constant adaptation, and low-pass-filter structure of
\citet{wu2025L1quad}.

Let $\widetilde{\mathbf v}=\widehat{\mathbf v}-\mathbf v$ and
$\widetilde{\boldsymbol\omega}=\widehat{\boldsymbol\omega}-\boldsymbol\omega$ be
the linear- and angular-velocity prediction errors. Stable diagonal predictor
matrices $\mathbf A_v$ and $\mathbf A_\omega$ provide error feedback in the two
state predictors. At each controller step $\Delta t$, the piecewise-constant
adaptation signals are
\begin{equation}
\begin{aligned}
\mathbf h_v
&=-\left(e^{\mathbf A_v\Delta t}-\mathbf I\right)^{-1}
\mathbf A_v e^{\mathbf A_v\Delta t}\widetilde{\mathbf v},\\
\mathbf h_\omega
&=-\left(e^{\mathbf A_\omega\Delta t}-\mathbf I\right)^{-1}
\mathbf A_\omega e^{\mathbf A_\omega\Delta t}\widetilde{\boldsymbol\omega}.
\end{aligned}
\end{equation}
These signals give the raw matched-uncertainty estimates
$\boldsymbol\sigma_F=m_{\mathrm{sys}}\mathbf R^\top\mathbf h_v$ and
$\boldsymbol\sigma_\tau=\mathbf J_{\mathrm{sys}}\mathbf h_\omega$, which are projected onto
configured Euclidean-norm bounds. Each projected estimate is passed through a
first-order low-pass filter,
\begin{equation}
\overline{\boldsymbol\sigma}_k
=\beta\overline{\boldsymbol\sigma}_{k-1}
+(1-\beta)\boldsymbol\sigma_k,
\qquad
\beta=e^{-2\pi f_c\Delta t},
\end{equation}
and the adaptive compensation is
$\mathbf F_{L_1}^b=-\overline{\boldsymbol\sigma}_F$ and
$\boldsymbol\tau_{L_1}^b=-\overline{\boldsymbol\sigma}_\tau$. The final command
adds these terms to the nominal force and torque. 

\subsubsection{Manipulator Position Tracking}

Neither base controller directly computes manipulator torques. The IK or direct
action pipeline supplies $\boldsymbol\theta_{a,d}$ to the simulator's implicit joint
position actuators. For each arm joint, the actuator applies
the PD relation $\tau_{a,j}=k_s(\theta_{a,d,j}-\theta_{a,j})-k_d\dot\theta_{a,j}$, subject to its
effort and velocity limits.

\subsubsection{Model Predictive Control}

For whole-body control, we formulate a nonlinear model predictive controller using the
\texttt{acados} framework \cite{verschueren2020acadosmodularopensourceframework}.
The MPC is solved with a prediction horizon of $H=32$ shooting intervals and a total
horizon duration of $t_{\mathrm{hor}}=0.8~\mathrm{s}$. The state and control inputs are defined as
\begin{align}
\mathbf{x}
&=
\begin{bmatrix}
\mathbf{p}^{\top} &
\mathbf{v}^{\top} &
\boldsymbol{\eta}^{\top} &
\boldsymbol{\omega}^{\top} &
\boldsymbol{\theta}_{a}^{\top}
\end{bmatrix}^{\top}
\in \mathbb{R}^{16},
\\
\mathbf{u}
&=
\begin{bmatrix}
({\mathbf{F}^{w}})^{\top} &
({\boldsymbol{\tau}^{w}})^{\top} &
\boldsymbol{\theta}_{a,\mathrm{ref}}^{\top}
\end{bmatrix}^{\top}
\in \mathbb{R}^{10}.
\end{align}
Here, $\mathbf{p}$ and $\mathbf{v}$ denote the position and linear velocity of the
aerial base, $\boldsymbol{\eta}$ and $\boldsymbol{\omega}$ denote the base Euler angles
and angular velocity, and $\boldsymbol{\theta}_{a}$ denotes the manipulator joint angles. The
control input consists of the commanded world-frame net base force $\mathbf{F}^{w}$, the commanded
world-frame base torque $\boldsymbol{\tau}^{w}$, and the manipulator joint reference
$\boldsymbol{\theta}_{a,\mathrm{ref}}$.

The MPC problem is formulated as
\begin{subequations}
\begin{align}
\mathbf{u}_{0:H-1}^{\star}
=
\arg\min_{\mathbf{u}_{0:H-1}}
\quad &
\ell_{e}(\mathbf{x}_{H})
+
\sum_{t=0}^{H-1}
\ell_{r}(\mathbf{x}_{t},\mathbf{u}_{t},\mathbf{u}_{t-1})
\\
\mathrm{s.t.}\quad
&
\mathbf{x}_{t+1}
=
\mathbf{f}_{\mathrm{dyn}}(\mathbf{x}_{t},\mathbf{u}_{t}),
\quad t=0,\ldots,H-1,
\\
&
\mathbf{x}_{0}=\hat{\mathbf{x}},
\qquad
\mathbf{x}_{t}\in\mathcal{X},
\\
&
\mathbf{u}_{\mathrm{lb}}
\leq
\mathbf{u}_{t}
\leq
\mathbf{u}_{\mathrm{ub}} .
\end{align}
\end{subequations}
The continuous-time dynamics used in the shooting model are
\begin{subequations}
\begin{align}
\dot{\mathbf{p}} &= \mathbf{v},
\\
\dot{\mathbf{v}} &= \frac{\mathbf{F}^{w}}{m_{\mathrm{sys}}},
\\
\dot{\boldsymbol{\eta}} &= \boldsymbol{\omega},
\\
\dot{\boldsymbol{\omega}}
&=
\begin{bmatrix}
\tau^{w}_{x}/J_{\mathrm{sys},xx} \\
\tau^{w}_{y}/J_{\mathrm{sys},yy} \\
\tau^{w}_{z}/J_{\mathrm{sys},zz}
\end{bmatrix},
\\
\dot{\boldsymbol{\theta}}_{a}
&=
k_a
\left(
\boldsymbol{\theta}_{a,\mathrm{ref}}
-
\boldsymbol{\theta}_{a}
\right).
\end{align}
\end{subequations}
Here, $m_{\mathrm{sys}}$ is the total system mass, $J_{\mathrm{sys},xx}$, $J_{\mathrm{sys},yy}$, and $J_{\mathrm{sys},zz}$ are the principal moments
of the aggregate system inertia, and $k_a$ is the first-order joint-response gain. Gravity compensation
is added to the optimized force command before applying the wrench to the aerial
platform. %

The stage cost is implemented as a nonlinear least-squares objective:
\begin{align}
\ell_r
=&\
\left\|
\mathbf{e}_{\mathrm{ee},t}
\right\|_{\mathbf{Q}_{\mathrm{ee}}}^2
+
\left\|
\mathbf{v}_{t}
\right\|_{\mathbf{Q}_{v}}^2
+
\left\|
\boldsymbol{\eta}_{t}
-
\boldsymbol{\eta}_{t}^{\mathrm{ref}}
\right\|_{\mathbf{Q}_{\eta}}^2
+
\left\|
\boldsymbol{\omega}_{t}
\right\|_{\mathbf{Q}_{\omega}}^2
\nonumber\\
&+
\left\|
\boldsymbol{\theta}_{a,t}
-
\boldsymbol{\theta}_{a,0}
\right\|_{\mathbf{Q}_{\theta}}^2
+
\left\|
\mathbf{u}_{t}
-
\mathbf{u}_{t}^{\mathrm{ref}}
\right\|_{\mathbf{W}_{u}}^2
+
\left\|
\mathbf{u}_{t}
-
\mathbf{u}_{t-1}
\right\|_{\mathbf{W}_{\Delta u}}^2,
\end{align}
where $\|\mathbf{a}\|_{\mathbf{M}}^2=\mathbf{a}^{\top}\mathbf{M}\mathbf{a}$ and
\begin{equation}
\mathbf{e}_{\mathrm{ee},t}
=
\begin{bmatrix}
\mathbf{p}_{\mathrm{ee}}(\mathbf{x}_{t})-\mathbf{p}_{\mathrm{ee},t}^{\mathrm{ref}} \\
\mathbf{e}_{R,\mathrm{ee}}^{\mathrm{MPC}}(\mathbf{x}_{t},\mathbf{R}_{\mathrm{ee},t}^{\mathrm{ref}})
\end{bmatrix}.
\end{equation}

The terms $\mathbf{p}_{\mathrm{ee}}(\mathbf{x}_{t})$ and
$\mathbf{e}_{R,\mathrm{ee}}^{\mathrm{MPC}}(\mathbf{x}_{t},\mathbf{R}_{\mathrm{ee},t}^{\mathrm{ref}})$ denote the end-effector
position and orientation residuals computed through the whole-body forward kinematics.
The orientation residual is represented in roll-pitch-yaw coordinates. The reference
base yaw is aligned with the desired end-effector yaw, while the reference roll and
pitch are set to zero to maintain stable flight. The nominal arm posture
$\boldsymbol{\theta}_{a,0}$ regularizes the manipulator configuration. Accordingly,
$\boldsymbol{\eta}_{t}^{\mathrm{ref}}$ contains zero reference roll and pitch and the desired end-effector yaw, while $\mathbf{u}_{t}^{\mathrm{ref}}$ contains zero net force and torque and the nominal arm posture. The matrices
$\mathbf{Q}_{\mathrm{ee}}$, $\mathbf{Q}_{v}$, $\mathbf{Q}_{\eta}$, $\mathbf{Q}_{\omega}$, and $\mathbf{Q}_{\theta}$ weight the tracking and state-regulation terms; $\mathbf{W}_{u}$ and $\mathbf{W}_{\Delta u}$ weight the control effort and control-rate penalty, respectively.
The terminal cost $\ell_e$ uses the same tracking and state-regulation residuals as
the stage cost but excludes the control-effort and control-rate terms.

The constraint set $\mathcal{X}$ includes bounds on the base position, base roll and
pitch, and manipulator joint angles. In addition, the relative end-effector height is
constrained as
\begin{equation}
0.1
\leq
p_{\mathrm{ee},z}(\mathbf{x}_{t}) - [\mathbf p_t]_z
\leq
2.5 .
\end{equation}
This constraint prevents the end-effector from moving below the aerial base and reduces
the risk of self-collision. 
The resulting
nonlinear optimal control problem is solved using an explicit Runge--Kutta integrator,
a Gauss--Newton Hessian approximation, a partial-condensing HPIPM QP solver, and SQP-RTI
iterations.

\subsection{Disturbances, Rotor Saturation, and Actuator Dynamics}
\label{app:disturbance-actuator}

We model three external disturbances: (1) aerodynamic drag and wind, (2) ground effect, and (3) near-wall effect. These models are combined with per-rotor thrust limits and optional transient actuator dynamics. After introducing the shared notation in Table~\ref{tab:nomenclature}, we describe the base-level and rotor-level disturbance models, their wrench reconstruction, and the actuator model that produces the thrust entering the proximity models.

\begin{table}[H]
\centering
\caption{Nomenclature for disturbance and rotor actuator models.}
\label{tab:nomenclature}
\small
\begin{tabularx}{\linewidth}{@{}lX@{}}
\hline
\textbf{Notation} & \textbf{Description} \\ \hline
$r_p$ & Rotor propeller radius \\
$z_i$ & Distance from the $i$-th rotor to the ground \\
$d_{\mathrm{wall},i}$, $\bar{d}_i$ & Wall distance and normalized wall distance $d_{\mathrm{wall},i}/r_p$ \\
$\mathbf e_3$, $\mathbf e_d$ & Vertical unit vector and wall-normal unit vector \\
$\mathbf F_{\mathrm{dw}}^b$, $\mathbf F_{\mathrm{wind}}^b$ & Combined drag-and-wind force and wind force \\
$\mathbf C_{\mathrm{drag}}$, $\mathbf v^b$ & Drag matrix and body-frame velocity \\
$\rho$, $\gamma_h$, $\gamma_v$ & Ground-effect intensity and near-wall scaling functions \\
$\mathbf T_{\mathrm{pre},i}$ & Realized post-actuator thrust vector entering the proximity models \\
$\mathbf T_{\mathrm{ge},i}$, $\mathbf T_{\mathrm{act},i}$ & Rotor thrust vectors after ground effect and after all proximity effects \\
$\mathbf T_{0,i}$, $\mathbf T_{h,i}$, $\mathbf T_{v,i}$ & Near-wall vertical input, horizontal component, and modified vertical component \\
$\mathbf A$, $\mathbf w_{\mathrm{prox}}^b$ & Control allocation matrix and body-frame proximity wrench \\
$\mathbf w_k^b$, $\mathbf T^{\mathrm{alloc}}_k$ & Desired body wrench and raw allocated rotor-thrust magnitudes at step $k$ \\
$T^{\mathrm{alloc}}_{i,k}$, $T^c_{i,k}$ & Raw allocated and thrust-limited magnitudes for rotor $i$ at step $k$ \\
$T^{\mathrm{dyn}}_{i,k}$, $\mathbf n_i$ & Realized pre-aerodynamic thrust magnitude and unit thrust direction \\
$T_{\min}$, $T_{\max}$ & Minimum and maximum per-rotor thrust \\
$q^c_{i,k}$, $q_{i,k}$ & Commanded and realized normalized rotor speeds \\
$\Delta t$, $\tau$, $\dot q_{\max}$ & Actuator timestep, response time constant, and normalized-speed rate cap in $\mathrm{s}^{-1}$ \\
$\lVert \cdot \rVert$ & Euclidean norm \\ \hline
\end{tabularx}
\end{table}

\subsubsection{Aerodynamic Drag and Wind}
Aerodynamic drag and wind are modeled as lumped forces acting on the multirotor base. Following the drag model in \cite{jacinto2023pegasus}, the combined force in the body frame is
\begin{equation*}
    \mathbf F_{\mathrm{dw}}^b = -\mathbf C_{\mathrm{drag}} \mathbf v^b + \mathbf F_{\mathrm{wind}}^b,
\end{equation*}
where $\mathbf C_{\mathrm{drag}} \in \mathbb{R}^{3 \times 3}$ is the diagonal drag coefficient matrix and $\mathbf F_{\mathrm{wind}}^b$ is a constant world-frame wind force transformed into the body frame.

\subsubsection{Ground Effect}
In contrast to base-level disturbances, proximity effects are modeled at the individual rotor level. This approach captures not only net force changes but also the resulting torque disturbances, particularly when the platform hovers partially over an asymmetric surface such as a desk edge. Let $\mathbf T_{\mathrm{ge},i}$ denote the output of the ground-effect stage. When ground effect is disabled, $\mathbf T_{\mathrm{ge},i}=\mathbf T_{\mathrm{pre},i}$; otherwise, the interaction with the ground is modeled as a thrust-augmentation effect \cite{kan2019analysis, danjun2015autonomous}:
\begin{equation*}
    \frac{\lVert \mathbf T_{\mathrm{pre},i} \rVert}{\lVert \mathbf T_{\mathrm{ge},i} \rVert} = 1 - \rho \left( \frac{r_p}{4z_i} \right)^2
\end{equation*}
where $\rho > 0$ is a constant parameter representing the ground-effect intensity.

\subsubsection{Near-Wall Effect}
The near-wall stage takes $\mathbf T_{\mathrm{ge},i}$ as its input, so ground effect is applied before near-wall effect when both are enabled. We adopt the model from \cite{ding2023aerodynamic}, which increases the magnitude of the component normal to the ground and adds a horizontal attraction force toward the wall. The final rotor thrust $\mathbf T_{\mathrm{act},i}$ is decomposed into a horizontal component $\mathbf T_{h,i}$ and a vertical component $\mathbf T_{v,i}$ as follows:
\begin{equation*}
\begin{aligned}
    \mathbf T_{0,i} &= (\mathbf T_{\mathrm{ge},i}^\top \mathbf e_3) \mathbf e_3, \\
    \mathbf T_{h,i} &= \gamma_h(\bar{d}_i) \lVert \mathbf T_{0,i} \rVert \mathbf e_d + (\mathbf T_{\mathrm{ge},i} - \mathbf T_{0,i}), \\
    \mathbf T_{v,i} &= \gamma_v(\bar{d}_i) \lVert \mathbf T_{0,i} \rVert \mathbf e_3, \\
    \mathbf T_{\mathrm{act},i} &= \mathbf T_{h,i} + \mathbf T_{v,i}.
\end{aligned}
\end{equation*}
where $\gamma_h(\bar{d}_i) = a_1 b_1^{\bar{d}_i}$ and $\gamma_v(\bar{d}_i) = 1 + a_2 b_2^{\bar{d}_i}$, with constant parameters $a_1, a_2 > 0$ and $b_1, b_2 \in (0,1)$.
When near-wall effect is disabled, $\mathbf T_{\mathrm{act},i}=\mathbf T_{\mathrm{ge},i}$.

\subsubsection{Proximity-Effect Wrench Reconstruction}
These rotor-level perturbations can be mapped to a body-wrench disturbance using the control allocation matrix $\mathbf A$ \cite{rashad2020fully}. The resulting lumped proximity wrench $\mathbf w_{\mathrm{prox}}^b$ in the body frame is
\begin{equation*}
    \mathbf w_{\mathrm{prox}}^b = \mathbf A \begin{bmatrix} \mathbf T_{\mathrm{act},1} - \mathbf T_{\mathrm{pre},1} \\ \vdots \\ \mathbf T_{\mathrm{act},N_r} - \mathbf T_{\mathrm{pre},N_r} \end{bmatrix},
\end{equation*}
where $N_r$ is the number of rotors and $\mathbf T_{\mathrm{act},i} - \mathbf T_{\mathrm{pre},i}$ represents the change in the thrust vector for the $i$-th rotor due to the proximity effects.

The distances $z_i$ and $d_{\mathrm{wall},i}$ are computed online using IsaacSim's raycasting. For ground distance, one ray is cast from each rotor opposite its current thrust axis, and the nearest non-robot hit gives $z_i$. For wall clearance, ten horizontal rays are cast from each rotor over a $90^\circ$ fan centered on the outward radial direction from the base center through that rotor. The nearest non-robot hit determines the wall direction $\mathbf e_d$; subtracting the propeller radius from its distance, with a lower bound of zero, gives $d_{\mathrm{wall},i}$ and $\bar d_i=d_{\mathrm{wall},i}/r_p$.

\subsubsection{Rotor Saturation and Reduced-Order Actuator Dynamics}
Given the desired body wrench $\mathbf w_k^b$, inverse control allocation computes the raw rotor-thrust magnitudes $\mathbf T^{\mathrm{alloc}}_k=\mathbf A_{\mathrm{mag}}^{\dagger}\mathbf w_k^b$, whose $i$-th entry is $T^{\mathrm{alloc}}_{i,k}$; see Section~\ref{app:control-allocation}. We first enforce the physical thrust range and convert the result to commanded normalized rotor speed:
\begin{equation}
    T^{c}_{i,k}
    =
    \operatorname{clip}\!\left(T^{\mathrm{alloc}}_{i,k},T_{\min},T_{\max}\right),
    \qquad
    q^{c}_{i,k}
    =
    \sqrt{\frac{T^{c}_{i,k}}{T_{\max}}},
    \label{eq:actuator-command}
\end{equation}

The first-order response is integrated exactly over one actuator timestep:
\begin{equation}
    \widetilde q_{i,k}
    =
    q_{i,k-1}
    +
    \left(1-e^{-\Delta t/\tau}\right)
    \left(q^{c}_{i,k}-q_{i,k-1}\right),
    \label{eq:actuator-first-order}
\end{equation}
A separate symmetric rate limit is then applied:
\begin{equation}
\begin{aligned}
    q'_{i,k}
    &=
    \operatorname{clip}\!\left(
        \widetilde q_{i,k},
        q_{i,k-1}-\dot q_{\max}\Delta t,
        q_{i,k-1}+\dot q_{\max}\Delta t
    \right),\\
    q_{i,k}
    &=
    \operatorname{clip}\!\left(
        q'_{i,k},
        \sqrt{T_{\min}/T_{\max}},
        1
    \right),
    \qquad
    T^{\mathrm{dyn}}_{i,k}=T_{\max}q_{i,k}^{2},
    \label{eq:actuator-rate-limit}
\end{aligned}
\end{equation}
Either transient effect can be disabled independently; disabling both gives $T^{\mathrm{dyn}}_{i,k}=T^c_{i,k}$. Suppressing the time index, the realized thrust vector entering the proximity models is $\mathbf T_{\mathrm{pre},i}=T^{\mathrm{dyn}}_{i}\mathbf n_i$.

\section{Experiment Details}
\label{app:additional-exp}

\subsection{Evaluation Metric Definitions}
\label{app:evaluation-metrics}
Let $N_t$ denote the number of timesteps in a rollout. Success rate is the fraction of all evaluation rollouts that satisfy the task success criterion in Table~\ref{tab:tasks}; subtask completion is likewise averaged over all evaluation rollouts. The diagnostic metrics---normalized end-effector and base tracking errors, tilt use, and rotor saturation rate---are computed within each rollout and averaged across successful evaluation rollouts. When uncertainty is shown, these metrics are reported as mean $\pm$ standard deviation across successful evaluation rollouts. When no successful rollout exists for a condition, these diagnostic metrics are reported as N/A.
A tracking metric is also reported as N/A when the corresponding commanded reference trajectory is not defined for that interface. Following~\citet{suarez2020benchmarks}, the tracking metrics $\rho_{\text{ee}}$ and $\rho_{\text{base}}$ are divided by the full arm reach $L_{\mathrm{arm}}$, making them dimensionless and comparable across robots of different sizes and morphologies.

\begin{itemize}[leftmargin=*]
    \item Subtask completion rate

    For a task with $K$ listed binary subtask criteria, let $c_k \in \{0,1\}$ indicate whether criterion $k$ is satisfied in a rollout. We define the subtask completion rate as
    \begin{equation*}
        r_{\mathrm{subtask}}
        =
        \frac{1}{K}
        \sum_{k=1}^{K}
        c_k .
    \end{equation*}
    Tasks whose subtask criteria are marked as \emph{None} are excluded from the subtask completion average.

    \item Normalized end-effector position tracking error (Norm. EE Error)
    
    Let $\mathbf{p}_{\mathrm{ee}}(t)$ and $\mathbf{p}^{\mathrm{ref}}_{\mathrm{ee}}(t)$ denote the measured and commanded end-effector positions at timestep $t$. We define
    \begin{equation*}
        \rho_{\text{ee}}
        =
        \frac{1}{N_t}
        \sum_{t=1}^{N_t}
        \frac{
        \left\|
        \mathbf{p}^{\mathrm{ref}}_{\mathrm{ee}}(t) - \mathbf{p}_{\mathrm{ee}}(t)
        \right\|_2
        }{
        L_{\mathrm{arm}}
        }.
    \end{equation*}

    \item Normalized base position tracking error (Norm. Base Error)
    
    Let $\mathbf{p}_{b}(t)$ and $\mathbf{p}^{\mathrm{ref}}_{b}(t)$ denote the measured and commanded base positions. We define the normalized base position tracking error as
    \begin{equation*}
        \rho_{\text{base}}
        =
        \frac{1}{N_t}
        \sum_{t=1}^{N_t}
        \frac{
        \left\|
        \mathbf{p}^{\mathrm{ref}}_{b}(t) - \mathbf{p}_{b}(t)
        \right\|_2
        }{
        L_{\mathrm{arm}}
        }.
    \end{equation*}

    \item Tilt use
    
    Let $\eta_{\mathrm{roll}}(t)$ and $\eta_{\mathrm{pitch}}(t)$ denote the base roll and pitch angles at timestep $t$. We define tilt use as the maximum roll-pitch tilt magnitude over the rollout:
    \begin{equation*}
        U_{\mathrm{tilt}}
        =
        \max_{t \in \{1,\dots,N_t\}}
        \sqrt{
        \eta_{\mathrm{roll}}(t)^2 + \eta_{\mathrm{pitch}}(t)^2
        }.
    \end{equation*}

    \item Rotor saturation rate (Sat. Rate)
    
    Let $\mathbf T_{\mathrm{pre},i}(t)$ be the pre-proximity thrust vector of rotor $i$ at timestep $t$ after inverse allocation, thrust limiting, and any enabled actuator dynamics. This is the rotor-thrust signal logged before the ground- and near-wall-effect stages. For the instantaneous-actuator setting used in the reported tracking studies, the saturation indicator is
    \begin{equation*}
        \chi^{\mathrm{sat}}_t
        =
        \mathbb{I}
        \left[
        \exists i \in \{1,\dots,N_r\}
        \ \text{s.t.}\
        \lVert \mathbf T_{\mathrm{pre},i}(t) \rVert \leq T_{\min}
        \ \lor\
        \lVert \mathbf T_{\mathrm{pre},i}(t) \rVert \geq T_{\max}
        \right],
        \end{equation*}
        and the saturation rate is
    \begin{equation*}
        r_{\mathrm{sat}}
        =
        \frac{1}{N_t}
        \sum_{t=1}^{N_t}
        \chi^{\mathrm{sat}}_t .
    \end{equation*}
\end{itemize}

\subsection{High-Level Visuomotor Policy Implementation Details}
\label{app:high-level-policy}

\subsubsection{Dataset and Common Policy Interface}
Each task-specific data-collection policy follows a predefined trajectory under randomized reset conditions and adds small perturbations to its waypoints. Only successful episodes are retained. We record $384\times384$ end-effector camera images, measured end-effector poses, gripper states, and absolute end-effector commands at $120$~Hz over a task-specific horizon. The resulting canonical dataset contains $80$ demonstrations per task; aggregating the $12$ task-specific datasets yields $960$ demonstrations for multi-task training.

All learned policies consume end-effector camera images and proprioceptive states and predict trajectories of end-effector poses and gripper commands. The recordings are downsampled by a factor of six to $20$~Hz for policy training and evaluation. Following UMI~\cite{chi2024universal}, pose trajectories are represented relative to the measured end-effector pose in the observation that begins each prediction chunk. 
\subsubsection{ACT}
ACT~\cite{zhao2023learning} is trained separately for each task. It uses relative action chunks. The visual backbone is initialized from ImageNet weights; images use ImageNet statistics, while states and actions are normalized using means and standard deviations computed from the complete task-specific training dataset. Additional parameters are shown in Table~\ref{tab:act-params}.

\subsubsection{Diffusion Policy}
DP~\cite{chi2023diffusionpolicy} is also trained separately for each task. We use a one-dimensional UNet denoiser conditioned on a pretrained CLIP visual encoder. Additional parameters are shown in Table~\ref{tab:dp-params}.

\subsubsection{$\pi_0$ and $\pi_{0.5}$}
We evaluate three adaptation levels; both adapted variants use full-parameter fine-tuning. \emph{Zero-shot} (ZS) applies the pretrained checkpoint without gradient updates. \emph{Multi-task fine-tuning} (MT-FT) initializes from the pretrained checkpoint and trains on the aggregated multi-task dataset. \emph{MT $\rightarrow$ ST-FT} initializes each task-specific model from the MT-FT checkpoint and continues training on that task's demonstrations.

Table~\ref{tab:policy-learning-params} summarizes the key hyperparameters used for training ACT, DP, $\pi_0$, and $\pi_{0.5}$.

\begin{table}[H]
\centering
\caption{Policy-learning hyperparameters.}
\label{tab:policy-learning-params}
\footnotesize
\setlength{\tabcolsep}{1.5pt}
\begin{subtable}[t]{0.30\textwidth}
\centering
\caption{ACT}
\label{tab:act-params}
\begin{tabular}{@{}lc@{}}
\toprule
\textbf{Parameter} & \textbf{Value} \\
\midrule
LR & $1 \times 10^{-5}$ \\
Batch size & 8 \\
Vision encoder & ResNet-18 \\
Obs horizon & 1 \\
Encoder layers & 4 \\
Decoder layers & 1 \\
FF dim & 3200 \\
Hidden dim & 512 \\
Attention heads & 8 \\
Chunk size & 16 \\
KL weight ($\beta$) & 10 \\
Train steps & 20000 \\
Save interval & 5000 \\
Optimizer & AdamW \\
Weight decay & $1\times10^{-4}$ \\
\bottomrule
\end{tabular}
\end{subtable}
\hspace{0.02\textwidth}
\begin{subtable}[t]{0.30\textwidth}
\centering
\caption{Diffusion Policy}
\label{tab:dp-params}
\begin{tabular}{@{}lc@{}}
\toprule
\textbf{Parameter} & \textbf{Value} \\
\midrule
Image obs horizon & 2 \\
Proprio obs horizon & 2 \\
Action horizon & 16 \\
Image resolution & $224 \times 224$ \\
Vision encoder & ViT-Base/16 \\
DDIM steps & 16 \\
LR & $3.0 \times 10^{-4}$ \\
Train epochs & 40 \\
Batch size & 64 \\
Validation split & $5\%$ \\
LR schedule & Cosine \\
Warmup steps & 2000 \\
\bottomrule
\end{tabular}
\end{subtable}
\hspace{0.02\textwidth}
\begin{subtable}[t]{0.30\textwidth}
\centering
\caption{$\pi_0$, $\pi_{0.5}$}
\label{tab:pi-params}
\begin{tabular}{@{}lc@{}}
\toprule
\textbf{Parameter} & \textbf{Value} \\
\midrule
Action horizon & 50 \\
MT-FT steps & 40000 \\
ST-FT steps & 20000 \\
MT batch size & 32 \\
ST batch size & 8 \\
Save interval & 5000 \\
Peak LR & $2.5\times10^{-5}$ \\
Warmup steps & 1000 \\
Optimizer & AdamW \\
EMA decay & 0.99 \\
\bottomrule
\end{tabular}
\end{subtable}
\end{table}

\section{Theoretical Background for AM Dynamics}
\label{app:theory-am-dynamics}

We discuss the dynamic characteristics that distinguish aerial manipulators from tabletop or ground-based mobile manipulators. In particular, we focus on the coupled dynamics between the multirotor base and manipulator arm and on the end-effector tracking challenges arising from the underactuated nature of the multirotor base.%

\subsection{Dynamic Coupling Between the Multirotor Base and Manipulator}
\label{app:dynamics-coupling}

The system variables, control inputs, and physical parameters are summarized in Table~\ref{table:system_notation}. Based on these definitions, the configuration $r$ and the combined velocity $\mathbf v_r$ describe the coupled rotational and joint motions. For conventional underactuated multirotor bases, the body-frame base force is further restricted to $\mathbf F^b = T_{\mathrm{tot}} \mathbf e_3$, where $T_{\mathrm{tot}} \in \mathbb{R}$ denotes the total thrust magnitude.

\begin{table}[H]
\centering
\caption{System configuration and physical parameters.}
\label{table:system_notation}
\begin{tabular}{ll}
\hline
\textbf{Notation} & \textbf{Description} \\ \hline
$\mathbf p \in \mathbb{R}^3$ & Base position \\
$\mathbf R \in SO(3)$ & Base orientation \\
$\boldsymbol\omega \in \mathbb{R}^3$ & Base angular velocity (body frame) \\
$\boldsymbol\theta_a \in \mathbb{R}^{n_a}$ & Manipulator joint angles \\
$r = (\mathbf R, \boldsymbol\theta_a) \in SO(3) \times \mathbb{R}^{n_a}$ & Combined rotational configuration \\
$\mathbf v_r = [\boldsymbol\omega^\top, \dot{\boldsymbol\theta}_a^\top]^\top \in \mathbb{R}^{3+n_a}$ & Combined angular/joint velocity \\
$\mathbf F^b \in \mathbb{R}^3$ & Base force (body frame) \\
$\boldsymbol\tau^b \in \mathbb{R}^3$ & Base torque (body frame) \\
$\boldsymbol\tau_a \in \mathbb{R}^{n_a}$ & Manipulator joint torques \\
$m_b, \mathbf J_b$ & Base mass and inertia \\
$m_{\mathrm{sys}}$ & Total system mass \\
$g_0$ & Gravitational acceleration magnitude \\
\hline
\end{tabular}
\end{table}

\subsubsection{Dynamics Without a Manipulator}
For a standard multirotor without a manipulator, the translational and rotational open-loop dynamics are fully decoupled as follows:
\begin{subequations} \label{eq:wo_manip}
\begin{align}
    m_b \ddot{\mathbf p} &= \mathbf R \mathbf F^b - m_b g_0 \mathbf e_3, \label{eq:wo_manip_trans} \\
    \mathbf J_b \dot{\boldsymbol\omega} &= -\boldsymbol\omega \times \mathbf J_b \boldsymbol\omega + \boldsymbol\tau^b \label{eq:wo_manip_rot}
\end{align}
\end{subequations}
where $\mathbf e_3 = [0,0,1]^\top$.

\subsubsection{Dynamics With a Manipulator}
In contrast, the dynamics of an aerial manipulator exhibit coupling. Let $\mathbf p_c$ denote the center of mass (CoM) of the entire system. Following the formulation in \cite{kim2023globally}, the coupled dynamics are
\begin{subequations}
\begin{align}
    m_{\mathrm{sys}} \ddot{\mathbf p}_c &= \mathbf R \mathbf F^b - m_{\mathrm{sys}} g_0 \mathbf e_3 \label{eq:w_manip_trans} \\
    \begin{bmatrix}
        \mathbf M_b & \mathbf M_{b,a} \\
        \mathbf M_{b,a}^\top & \mathbf M_a
    \end{bmatrix} \dot{\mathbf v}_r &= \begin{bmatrix} \boldsymbol\tau^b \\ \boldsymbol\tau_a \end{bmatrix} + \mathbf c_r(\mathbf v_r, r, \dot{\mathbf p}_c, \mathbf F^b) \label{eq:w_manip_rot}
\end{align}
\end{subequations}
where $\mathbf M_b$, $\mathbf M_a$, and $\mathbf M_{b,a}$ are the sub-blocks of the generalized mass matrix, and $\mathbf c_r(\mathbf v_r, r, \dot{\mathbf p}_c, \mathbf F^b)$ includes gyroscopic torques and coupling effects involving the CoM velocity and force. 

These equations reveal the following coupling features:
\begin{itemize}
    \item The non-diagonal block $\mathbf M_{b,a}$ in \eqref{eq:w_manip_rot} indicates that the rotational motion of the base and the joint motions of the manipulator are inherently coupled.
    \item While the CoM translational dynamics \eqref{eq:w_manip_trans} appear decoupled in form, the rotational dynamics \eqref{eq:w_manip_rot} remain dependent on the translational state through the nonlinear term $\mathbf c_r(\mathbf v_r, r, \dot{\mathbf p}_c, \mathbf F^b)$. The base torque must compensate for this coupling term, placing additional demands on the multirotor's control system.
    \item The end-effector position is $\mathbf p_{\mathrm{ee}} = \mathbf p_c + \mathbf d(r)$, where $\mathbf d(r)$ is the displacement vector. This relationship makes end-effector tracking sensitive to the accuracy of both translational and rotational control.
\end{itemize}

\subsection{Effects of Underactuated Dynamics on End-Effector Tracking}
\label{app:underactuated-dynamics}

In conventional multirotor systems, inherent underactuation prevents independent control of the base position and its full orientation. For an underactuated multirotor, the body-frame force is constrained to $\mathbf F^b = T_{\mathrm{tot}} \mathbf e_3$. This constraint implies that the system lacks the authority to directly actuate all three translational DoFs otal thrust $T_{\mathrm{tot}} \in \mathbb{R}$ alone. To address this limitation, the translational dynamics can be reformulated as
\begin{equation*}
    m_b\ddot{\mathbf p} = \mathbf R_d T_{\mathrm{tot}} \mathbf e_3 - m_b g_0 \mathbf e_3 + (\mathbf R-\mathbf R_d) T_{\mathrm{tot}} \mathbf e_3, \label{eq:virtual_input_reformulation}
\end{equation*}
where $\mathbf R_d T_{\mathrm{tot}} \mathbf e_3$ is treated as a 3D virtual control input. This formulation enables full translational control, provided that the actual orientation $\mathbf R$ converges rapidly to the desired orientation $\mathbf R_d$. Based on this approach, a common choice is a cascaded control structure in which the outer-loop position controller computes the required $\mathbf R_d$ and $T_{\mathrm{tot}}$, while the inner-loop attitude controller ensures that $\mathbf R$ tracks $\mathbf R_d$ sufficiently fast \cite{lee2010geometric}. This fundamental characteristic remains a primary constraint for aerial manipulators with underactuated multirotor bases \cite{kim2013aerial}.

Consequently, for an aerial manipulator with an underactuated base, the target orientation $\mathbf R_d$ cannot be chosen arbitrarily because it is determined by the requirements of the position controller. The actual orientation $\mathbf R$ is governed by the closed-loop dynamics of the attitude controller. Furthermore, the end-effector position $\mathbf p_{\mathrm{ee}}(\mathbf p, \mathbf R, \boldsymbol\theta_a)$ and orientation $\mathbf R_{\mathrm{ee}}(\mathbf R, \boldsymbol\theta_a)$ depend kinematically on the base pose $(\mathbf p, \mathbf R)$ and joint angles $\boldsymbol\theta_a$, making precise task-space tracking inherently difficult. High-fidelity tracking therefore requires a controller that explicitly accounts for the coupling between the closed-loop translational and rotational dynamics.

\subsection{Control Allocation and Rotor Saturation}
\label{app:control-allocation}

We describe the control allocation process for the FA-Hexa platform. The same fixed-thrust-direction formulation applies to UA-Hexa and UA-Quad with their respective allocation matrices. For Omni-Hexa, the allocation is extended to jointly compute rotor thrust magnitudes and thrust-vectoring angles \cite{lee2025autonomous}.

Let $\mathbf T_1, \mathbf T_2, \dots, \mathbf T_6 \in \mathbb{R}^3$ denote the rotor thrust vectors. At timestep $k$, the low-level controller produces the desired body-frame wrench $\mathbf w_k^b=[({\mathbf F_k^b})^\top,({\boldsymbol{\tau}_k^b})^\top]^\top \in \mathbb{R}^6$. Control allocation determines the corresponding raw rotor-thrust magnitudes $\mathbf T^{\mathrm{alloc}}_k=[T^{\mathrm{alloc}}_{1,k},\dots,T^{\mathrm{alloc}}_{6,k}]^\top \in \mathbb{R}^6$. Based on the rotor geometry \cite{rashad2020fully}, the forward and inverse mappings are
\begin{equation*}
    \mathbf w_k^b=\mathbf A_{\mathrm{mag}}\mathbf T^{\mathrm{alloc}}_k,
    \qquad
    \mathbf T^{\mathrm{alloc}}_k=\mathbf A_{\mathrm{mag}}^\dagger\mathbf w_k^b,
\end{equation*}
where $\mathbf A_{\mathrm{mag}}^\dagger$ is the pseudoinverse, and the $i$-th column of the allocation matrix $\mathbf A_{\mathrm{mag}} \in \mathbb{R}^{6 \times 6}$ is
\begin{equation*}
    \mathbf A_{\mathrm{mag},i} = \begin{bmatrix} \mathbf n_i \\ \mathbf d_i \times \mathbf n_i + k_\tau (-1)^{\alpha_i} \mathbf n_i \end{bmatrix}.
\end{equation*}
Here, $\mathbf d_i \in \mathbb{R}^3$ is the displacement vector from the platform's CoM to the $i$-th rotor, $\mathbf n_i \in \mathbb{R}^3$ is the unit thrust direction vector, $\alpha_i \in \{0, 1\}$ denotes the direction of rotation, and $k_\tau$ is the reaction-moment-to-thrust ratio. The actuator model subsequently enforces $T_{\min}\leq T^c_{i,k}\leq T_{\max}$ as described in Section~\ref{app:disturbance-actuator}.

To account for possible changes in thrust direction, we further characterize the allocation problem using the individual rotor thrust vectors $\mathbf T_i \in \mathbb{R}^3$. Defining the collective thrust vector $\mathbf T_{\mathrm{col}} = [\mathbf T_1^\top, \dots, \mathbf T_6^\top]^\top \in \mathbb{R}^{18}$ yields
\begin{equation*}
    \mathbf w_k^b = \mathbf A \mathbf T_{\mathrm{col}},
\end{equation*}
where $\mathbf A \in \mathbb{R}^{6 \times 18}$ is composed of block matrices $\mathbf A = [\mathbf A_1, \dots, \mathbf A_6]$. The block $\mathbf A_i \in \mathbb{R}^{6 \times 3}$ corresponding to the $i$-th rotor is
\begin{equation*}
    \mathbf A_i = \begin{bmatrix} \mathbf I_3 \\ \widehat{\mathbf d}_i + k_\tau (-1)^{\alpha_i} \mathbf I_3 \end{bmatrix},
\end{equation*}
where $\mathbf I_3$ is the $3 \times 3$ identity matrix and $\widehat{\mathbf d}_i$ is the skew-symmetric matrix such that $\widehat{\mathbf d}_i \mathbf x = \mathbf d_i \times \mathbf x$ for all $ \mathbf x \in \mathbb{R}^3$.

\section{Additional Experiment Results}

\subsection{High-Level Visuomotor Policy}
\label{app:high-level-policy-results}
We report the per-task success rates and subtask completion rates for all evaluated visuomotor policies in Tables~\ref{tab:app-policy-success-full} and~\ref{tab:app-policy-subtask-full}.

\begin{table}[H]
\centering
\caption{Full per-task success rates for the high-level policy comparison. Each value is computed from 30 evaluation rollouts.}
\label{tab:app-policy-success-full}
\scriptsize
\setlength{\tabcolsep}{2.2pt}
\renewcommand{\arraystretch}{1.08}
\begin{tabular*}{\textwidth}{@{\extracolsep{\fill}}lcccccccc@{}}
\toprule
\multirow{2}{*}{\raisebox{-2.0ex}{Task}} &
\multirow{2}{*}{\raisebox{-2.0ex}{ACT}} &
\multirow{2}{*}{\raisebox{-2.0ex}{DP}} &
\multicolumn{3}{c}{\(\pi_0\)} &
\multicolumn{3}{c}{\(\pi_{0.5}\)} \\
\cmidrule(lr){4-6}
\cmidrule(lr){7-9}
& & &
ZS &
MT-FT &
\makecell{MT \(\to\) \\ ST-FT} &
ZS &
MT-FT &
\makecell{MT \(\to\) \\ ST-FT} \\
\midrule
\multicolumn{9}{l}{\textit{Instantaneous interaction}} \\
Press button & 80 & 73.3 & 33.3 & 93.3 & 56.7 & 0 & 100 & 96.7 \\
Peg-in-hole & 40 & 66.7 & 0 & 60 & 66.7 & 0 & 100 & 90 \\
\midrule
\multicolumn{9}{l}{\textit{Object transport}} \\
Frame assembly & 0 & 0 & 0 & 3.3 & 0 & 0 & 0 & 0 \\
Cabinet pick-and-place & 10 & 3.3 & 0 & 0 & 0 & 0 & 0 & 0 \\
Lemon harvesting & 3.3 & 26.7 & 0 & 16.7 & 50 & 0 & 66.7 & 86.7 \\
Toss ball & 0 & 0 & 0 & 0 & 16.7 & 0 & 0 & 10 \\
\midrule
\multicolumn{9}{l}{\textit{Articulated object and constrained contact}} \\
Rotate valve & 13.3 & 26.7 & 0 & 20 & 0 & 0 & 0 & 0 \\
Push slider & 23.3 & 63.3 & 0 & 20 & 100 & 0 & 100 & 90 \\
Pull lever & 13.3 & 83.3 & 20 & 76.7 & 90 & 0 & 66.7 & 83.3 \\
Open door & 100 & 93.3 & 50 & 93.3 & 90 & 33.3 & 100 & 100 \\
Wipe window & 0 & 3.3 & 0 & 0 & 0 & 0 & 0 & 10 \\
NDT & 40 & 56.7 & 0 & 56.7 & 70 & 0 & 56.7 & 60 \\
\midrule
\textbf{Macro average} & \textbf{26.94} & \textbf{41.39} & \textbf{8.61} & \textbf{36.67} & \textbf{45.00} & \textbf{2.78} & \textbf{49.17} & \textbf{52.22} \\
\bottomrule
\end{tabular*}
\end{table}

\begin{table}[H]
\centering
\caption{Full per-task subtask completion rates for the high-level policy comparison. Each value is computed from 30 evaluation rollouts; rows for tasks without defined subtask criteria are omitted.}
\label{tab:app-policy-subtask-full}
\scriptsize
\setlength{\tabcolsep}{2.2pt}
\renewcommand{\arraystretch}{1.08}
\begin{tabular*}{\textwidth}{@{\extracolsep{\fill}}lcccccccc@{}}
\toprule
\multirow{2}{*}{\raisebox{-2.0ex}{Task}} &
\multirow{2}{*}{\raisebox{-2.0ex}{ACT}} &
\multirow{2}{*}{\raisebox{-2.0ex}{DP}} &
\multicolumn{3}{c}{\(\pi_0\)} &
\multicolumn{3}{c}{\(\pi_{0.5}\)} \\
\cmidrule(lr){4-6}
\cmidrule(lr){7-9}
& & &
ZS &
MT-FT &
\makecell{MT \(\to\) \\ ST-FT} &
ZS &
MT-FT &
\makecell{MT \(\to\) \\ ST-FT} \\
\midrule
\multicolumn{9}{l}{\textit{Object transport}} \\
Frame assembly & 5 & 20 & 0 & 25 & 0 & 0 & 46.7 & 50 \\
Cabinet pick-and-place & 20 & 13.3 & 2.2 & 26.7 & 32.2 & 0 & 30 & 30 \\
Lemon harvesting & 5 & 43.3 & 0 & 60 & 80 & 0 & 73.3 & 88.3 \\
Toss ball & 50 & 50 & 0 & 50 & 53.3 & 23.3 & 38.3 & 58.3 \\
\midrule
\multicolumn{9}{l}{\textit{Articulated object and constrained contact}} \\
Rotate valve & 50 & 70 & 6.7 & 60 & 50 & 0 & 50 & 50 \\
Push slider & 45 & 80 & 0 & 60 & 100 & 0 & 100 & 96.7 \\
Pull lever & 15 & 83.3 & 20 & 80 & 90 & 0 & 66.7 & 95 \\
Open door & 100 & 93.3 & 50 & 96.7 & 96.7 & 33.3 & 100 & 100 \\
Wipe window & 17.5 & 28.3 & 1.7 & 33.3 & 29.2 & 0 & 30 & 37.5 \\
\midrule
\textbf{Macro average} & \textbf{34.17} & \textbf{53.52} & \textbf{8.95} & \textbf{54.63} & \textbf{59.04} & \textbf{6.30} & \textbf{59.44} & \textbf{67.31} \\
\bottomrule
\end{tabular*}
\end{table}

\subsubsection{Diffusion Policy Data Scaling}
\label{app:dp-data-scaling}

The amount of demonstration data required to learn effective policies is an important practical consideration. We evaluate how DP success rates scale with $10$, $20$, $40$, and $80$ demonstrations per task. Each trained policy is evaluated on $30$ evaluation rollouts.

\begin{table}[H]
\centering
\caption{Per-task success rates for DP trained with different numbers of demonstrations. Each value is computed from 30 evaluation rollouts.}
\label{tab:dp-data-scaling}
\scriptsize
\setlength{\tabcolsep}{2.2pt}
\renewcommand{\arraystretch}{1.08}
\begin{tabular*}{\textwidth}{@{\hspace{4pt}\extracolsep{\fill}}lcccc@{\hspace{4pt}}}
\toprule
Task & 10 demos & 20 demos & 40 demos & 80 demos \\
\midrule
\multicolumn{5}{l}{\textit{Instantaneous interaction}} \\
Press button & 33.3 & 63.3 & 86.7 & 86.7 \\
Peg-in-hole & 40 & 60 & 63.3 & 76.7 \\
\midrule
\multicolumn{5}{l}{\textit{Object transport}} \\
Frame assembly & 0 & 0 & 0 & 0 \\
Cabinet pick-and-place & 0 & 0 & 0 & 0 \\
Lemon harvesting & 6.7 & 40 & 36.7 & 56.7 \\
Toss ball & 0 & 6.7 & 16.7 & 16.7 \\
\midrule
\multicolumn{5}{l}{\textit{Articulated object and constrained contact}} \\
Rotate valve & 0 & 3.3 & 0 & 3.3 \\
Push slider & 30 & 83.3 & 100 & 93.3 \\
Pull lever & 36.7 & 76.7 & 53.3 & 50 \\
Open door & 96.7 & 100 & 100 & 100 \\
Wipe window & 3.3 & 6.7 & 0 & 6.7 \\
NDT & 0 & 0 & 13.3 & 56.7 \\
\midrule
\textbf{Macro average} & \textbf{20.6} & \textbf{36.7} & \textbf{39.2} & \textbf{45.6} \\
\bottomrule
\end{tabular*}
\end{table}

Macro-average success increases with the number of demonstrations, although the effect varies across tasks. Press button, peg-in-hole, lemon harvesting, and NDT benefit substantially from additional data, while pull lever and push slider show non-monotonic trends. Overall, $80$ demonstrations provide the strongest aggregate performance among the evaluated budgets.

\subsection{Policy--Control Interface}
\label{app:policy-control-results}

In addition to the DP results presented in the main text, this section provides the full policy--control interface results for ACT and \(\pi_{0.5}\) MT-FT.
\begin{table}[!ht]
  \centering
  \caption{ACT low-level interface ablation and mechanism breakdown. Each configuration is evaluated on 30 evaluation rollouts; diagnostic metrics are reported as mean $\pm$ standard deviation over successful evaluation rollouts.}
  \label{tab:exp2_act_mechanism_main}
  \scriptsize
  \setlength{\tabcolsep}{2.4pt}
  \renewcommand{\arraystretch}{1.12}
  \begin{tabular*}{\textwidth}{@{\extracolsep{\fill}}llccccc@{}}
    \toprule[0.8pt]
    & &
    \multicolumn{1}{c}{\textbf{Outcome}} &
    \multicolumn{2}{c}{\textbf{Tracking}} &
    \multicolumn{2}{c}{\textbf{Limit Use}} \\
    \cmidrule[0.1pt](lr){3-3}
    \cmidrule[0.1pt](lr){4-5}
    \cmidrule[0.1pt](l){6-7}
    \textbf{Task} &
    \textbf{Control Interface} &
    \textbf{SR [\%]  \(\uparrow\)} &
    \makecell{\textbf{Norm. EE} \textbf{Error [$10^{-2}$]} \(\downarrow\)} &
    \makecell{\textbf{Norm. Base} \textbf{Error [$10^{-2}$]} \(\downarrow\)} &
    \makecell{\textbf{Tilt Use}  \textbf{[rad]} \(\downarrow\)} &
    \makecell{\textbf{Sat.} \textbf{Rate [\%]} \(\downarrow\)} \\
    \midrule[0.7pt]
    \multirow{5}{*}[-1ex]{\makecell[l]{Press\\button}}
    & EE target, IK-PID & \textbf{80} & $0.481 \pm 0.227$ & $0.171 \pm 0.043$ & $\mathbf{0.023 \pm 0.007}$ & $12.44 \pm 1.82$\\
    & \makecell[l]{Base and arm joint\\target, PID} & 6.7 & N/A & $\mathbf{0.089 \pm 0.021}$ & $0.089 \pm 0.033$ & $\mathbf{10.04 \pm 0.64}$ \\
    \cmidrule[0.1pt]{2-7}
    & EE target, IK-$L_1$ & \textbf{66.7} & $0.248 \pm 0.005$ & $0.139 \pm 0.002$ & $\mathbf{0.014 \pm 0.000}$ & $\mathbf{10.65 \pm 0.23}$ \\
    & \makecell[l]{Base and arm joint\\target, $L_1$} & 13.3 & N/A & $\mathbf{0.121 \pm 0.006}$ & $0.100 \pm 0.002$ & $10.75 \pm 0.47$ \\
    \cmidrule[0.1pt]{2-7}
    & EE target, MPC & 83.3 & $1.295 \pm 0.013$ & N/A & $0.019 \pm 0.000$ & $0.01 \pm 0.01$\\
    \midrule[0.7pt]
    \multirow{5}{*}[-1ex]{\makecell[l]{Push\\slider}}
    & EE target, IK-PID & \textbf{23.3} & $ 0.879 \pm 0.107$ & $0.463 \pm 0.104$ & $0.188 \pm 0.053$ & $48.76 \pm 7.80$\\
    & \makecell[l]{Base and arm joint\\target, PID} & 0.0 & N/A & N/A & N/A & N/A \\
    \cmidrule[0.1pt]{2-7}
    & EE target, IK-$L_1$ & \textbf{23.3} & $1.080 \pm 0.204$ & $0.438 \pm 0.118$ & $0.041 \pm 0.007$ & $43.63 \pm 4.97$\\
    & \makecell[l]{Base and arm joint\\target, $L_1$} & 0.0 & N/A & N/A & N/A & N/A \\
    \cmidrule[0.1pt]{2-7}
    & EE target, MPC & 20.0 & $5.892 \pm 0.721$ & N/A & $0.031 \pm 0.001$ & $5.34 \pm 5.82$\\
    \bottomrule[0.8pt]
  \end{tabular*}
  \vspace{-0.25in}
\end{table}

\begin{table}[!ht]
  \centering
  \caption{$\pi_{0.5}$ MT-FT low-level interface ablation and mechanism breakdown. Each configuration is evaluated on 30 evaluation rollouts; diagnostic metrics are reported as mean $\pm$ standard deviation over successful evaluation rollouts.}
  \label{tab:pi_mechanism_main}
  \scriptsize
  \setlength{\tabcolsep}{2.4pt}
  \renewcommand{\arraystretch}{1.12}
  \begin{tabular*}{\textwidth}{@{\extracolsep{\fill}}llccccc@{}}
    \toprule[0.8pt]
    & &
    \multicolumn{1}{c}{\textbf{Outcome}} &
    \multicolumn{2}{c}{\textbf{Tracking}} &
    \multicolumn{2}{c}{\textbf{Limit Use}} \\
    \cmidrule[0.1pt](lr){3-3}
    \cmidrule[0.1pt](lr){4-5}
    \cmidrule[0.1pt](l){6-7}
    \textbf{Task} &
    \textbf{Control Interface} &
    \textbf{SR [\%]  \(\uparrow\)} &
    \makecell{\textbf{Norm. EE} \textbf{Error [$10^{-2}$]} \(\downarrow\)} &
    \makecell{\textbf{Norm. Base} \textbf{Error [$10^{-2}$]} \(\downarrow\)} &
    \makecell{\textbf{Tilt Use}  \textbf{[rad]} \(\downarrow\)} &
    \makecell{\textbf{Sat.} \textbf{Rate [\%]} \(\downarrow\)} \\
    \midrule[0.7pt]
    \multirow{5}{*}[-1ex]{\makecell[l]{Press\\button}}
    & EE target, IK-PID & \textbf{100.0} & $0.385 \pm 0.220$ & $0.175 \pm 0.064$ & $\mathbf{0.015 \pm 0.005}$ & $\mathbf{11.73 \pm 2.54}$\\
    & \makecell[l]{Base and arm joint\\target, PID} & 70.0 & N/A & $\mathbf{0.164 \pm 0.039}$ & $0.068 \pm 0.006$ & $21.22 \pm 5.15$ \\
    \cmidrule[0.1pt]{2-7}
    & EE target, IK-$L_1$ & \textbf{100.0} & $0.439 \pm 0.191$ & $\mathbf{0.167 \pm 0.044}$ & $\mathbf{0.014 \pm 0.012}$ & $\mathbf{22.20 \pm 10.20}$ \\
    & \makecell[l]{Base and arm joint\\target, $L_1$} & 30.0 & N/A & $0.178 \pm 0.106$ & $0.226 \pm 0.235$ & $33.13 \pm 1.60$ \\
    \cmidrule[0.1pt]{2-7}
    & EE target, MPC & 93.3 & $1.332 \pm 0.308$ & N/A & $0.031 \pm 0.021$ & $2.24 \pm 3.98$\\
    \midrule[0.7pt]
    \multirow{5}{*}[-1ex]{\makecell[l]{Push\\slider}}
    & EE target, IK-PID & \textbf{100.0} & $1.350 \pm 0.362$ & $\mathbf{0.512 \pm 0.160}$ & $\mathbf{0.033 \pm 0.017}$ & $\mathbf{16.92 \pm 2.64}$\\
    & \makecell[l]{Base and arm joint\\target, PID} & 43.3 & N/A & $1.514 \pm 0.494$ & $0.364 \pm 0.105$ & $42.86 \pm 4.65$ \\
    \cmidrule[0.1pt]{2-7}
    & EE target, IK-$L_1$ & \textbf{90.0} & $1.070 \pm 0.478$ & $\mathbf{0.652 \pm 0.139}$ & $\mathbf{0.085 \pm 0.021}$ & $\mathbf{66.73 \pm 6.87}$ \\
    & \makecell[l]{Base and arm joint\\target, $L_1$} & 60.0 & N/A & $0.820 \pm 0.239$ & $0.494 \pm 0.209$ & $68.05 \pm 15.91$ \\
    \cmidrule[0.1pt]{2-7}
    & EE target, MPC & 100.0 & $3.177 \pm 1.627$ & N/A & $0.056 \pm 0.009$ & $3.46 \pm 1.09$\\
    \bottomrule[0.8pt]
  \end{tabular*}
  \vspace{-0.25in}
\end{table}

\subsection{Actuator Dynamics Identification and Vertical Tracking}
\label{app:actuator-dynamics-experiment}

We identify the effective rotor response of the physical FA-Hexa from real-world flight-command and proprioceptive logs, following \citet{eschmann2024datadrivenidentificationquadrotorssubject}. The identified time constant is $\tau=47.5\,\mathrm{ms}$. Because the data do not reveal a repeatable hard plateau in normalized-speed acceleration, we adopt a normalized-speed rate cap of $\dot q_{\max}=0.75\,\mathrm{s}^{-1}$.

We evaluate the model by replaying two vertical sum-of-sinusoids references on FA-Hexa while holding the controller, embodiment, and other simulation settings fixed. The low-frequency reference spans $0.18$--$0.71\,\mathrm{Hz}$, and the high-frequency reference spans $0.25$--$1.50\,\mathrm{Hz}$. We compare four configurations: no motor dynamics, motor delay modeled by the identified first-order response, the acceleration limit, and both effects together. Table~\ref{tab:actuator-dynamics-ablation} reports the base-$z$ tracking RMSE. For the high-frequency reference, the motor-delay model increases RMSE by $19\%$. No rotor reaches the $23\,\mathrm{N}$ thrust limit. 

\vspace{-1em}
\begin{table}[H]
    \centering
    \caption{FA-Hexa vertical tracking under different actuator configurations. Values are base-$z$ RMSEs in cm.}
    \label{tab:actuator-dynamics-ablation}
    \small
    \begin{tabular}{lcc}
        \toprule
        \textbf{Actuator configuration}
        & \textbf{Low-frequency RMSE (cm)}
        & \textbf{High-frequency RMSE (cm)} \\
        \midrule
        No motor dynamics & 0.14 & 1.96 \\
        Motor delay & 0.58 & 2.33 \\
        Acceleration limit ($0.75\,\mathrm{s}^{-1}$) & 0.14 & 1.95 \\
        Motor delay and acceleration limit & 0.58 & 2.33 \\
        \bottomrule
    \end{tabular}
\end{table}

\subsection{Real-World Experiment Details}
\label{app:real-world-experiments}

\subsubsection{Ground-Effect Comparison}
\label{app:real-world-ground-effect}

The physical experiment commands the end-effector vertical reference
\begin{equation*}
    z_{\mathrm{ref}}(t)
    =
    0.8 + 0.2\cos\!\left(\frac{2\pi t}{30}\right)\ \mathrm{m},
\end{equation*}
while holding the lateral end-effector position and orientation fixed. The reference is executed through the EE target interface, with IK generating the desired whole-body configuration and the PID controller tracking the resulting base trajectory. The ground-effect-enabled simulation uses intensity $\rho=3.4$ and propeller radius $r_p=0.152\,\mathrm{m}$. The values reported in Fig.~\ref{fig:realworld_groundeffect} are simulation-to-real EE-$z$ RMSEs over the two shaded altitude intervals.

\subsubsection{Lemon-Harvesting Learning Pipeline}
\label{app:real-world-lemon}

For the hardware experiment, we use FA-Hexa and collect real-world demonstrations with a UMI gripper~\cite{chi2024universal}. We train DP on these demonstrations using the same observation modalities as in simulation: the end-effector camera and proprioceptive state. During deployment, the policy predicts end-effector commands that are executed through the EE target interface with IK-PID control.

\subsubsection{Wrench-Limit Step Response}
\label{app:real-world-wrench-limit}

\begin{figure}[H]
    \centering
    \includegraphics[width=0.99\linewidth]{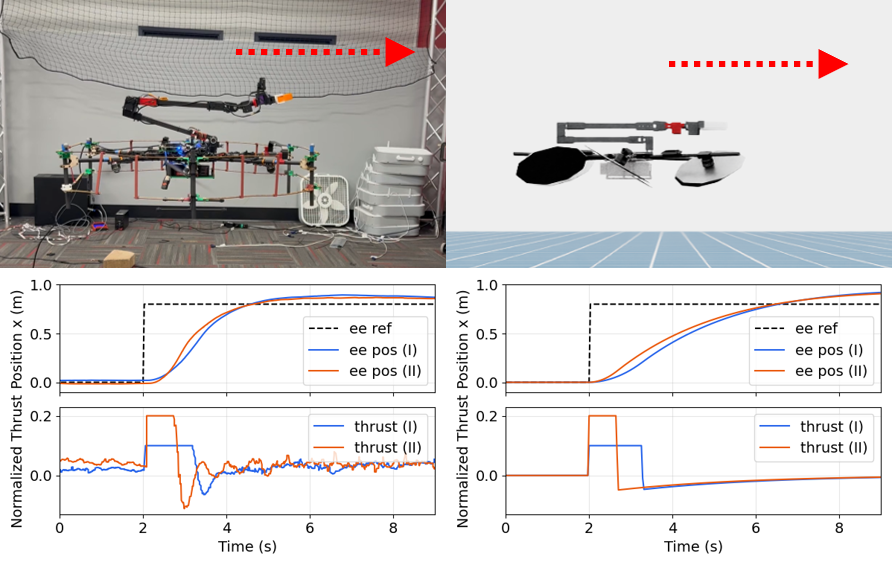}
    \caption{Real-world and simulated aerial-manipulator step responses under two wrench-space limits, denoted I and II.}
    \label{fig: realworld_stepx}
\end{figure}

We conduct additional experiments to assess whether simulation qualitatively reproduces real-world behavior under wrench-space limits. Specifically, we observe the $x$-axis step responses under two limit configurations. As shown in Fig.~\ref{fig: realworld_stepx}, the physical and simulated systems exhibit similar control-input saturation patterns when the limits are reached.

\end{document}